%% file: main.tex
\documentclass[lettersize,journal]{IEEEtran}
\usepackage{amsmath,amsfonts}
\usepackage{algorithmic}
\usepackage{algorithm}
\usepackage{array}
\usepackage[caption=false,font=normalsize,labelfont=sf,textfont=sf]{subfig}
\usepackage{textcomp}
\usepackage{stfloats}
\usepackage{url}
\usepackage{verbatim}
\usepackage{graphicx}
\usepackage{cite}
\usepackage{hyperref}
\usepackage{multirow, makecell}
\usepackage{booktabs}
\usepackage{color}
\begin{document}

\title{Robust Knowledge Adaptation for Dynamic Graph Neural Networks}

\author{Hanjie~Li,
Changsheng~Li,
Kaituo~Feng,
Ye~Yuan,
Guoren~Wang,
Hongyuan~Zha
\thanks{Hanjie Li, Changsheng Li, Kaituo Feng, Ye Yuan, and Guoren Wang are with the school of computer science and technology, Beijing Institute of Technology, Beijing, China. E-mail: \{lihanjieyouxiang@sina.com; lcs@bit.edu.cn; kaituofeng@gmail.com; yuan-ye@bit.edu.cn; wanggrbit@126.com.\}}
\thanks{Hongyuan Zha is with the School of Data Science, the Chinese University of Hong Kong, Shenzhen, China. E-mail: zhahy@cuhk.edu.cn.}
\thanks{Corresponding author: Changsheng Li.}
\thanks{This work was supported by the NSFC under Grants 62122013, U2001211. This work was also supported by the Innovative Development Joint Fund Key Projects of Shandong NSF under Grants ZR2022LZH007.}
}

\markboth{IEEE Transactions on Knowledge and Data Engineering}%
{Shell \MakeLowercase{\textit{et al.}}: A Sample Article Using IEEEtran.cls for IEEE Journals}


\maketitle

\input{sections/abstract}

\vspace{0.4in}
\input{sections/intro}
\input{sections/related_work}
\input{sections/method}
\input{sections/experiment}

\input{sections/conclusion}

\bibliographystyle{IEEEtran}
\bibliography{reference}


\begin{IEEEbiography}[{\includegraphics[width=1in,height=1.25in,clip,keepaspectratio]{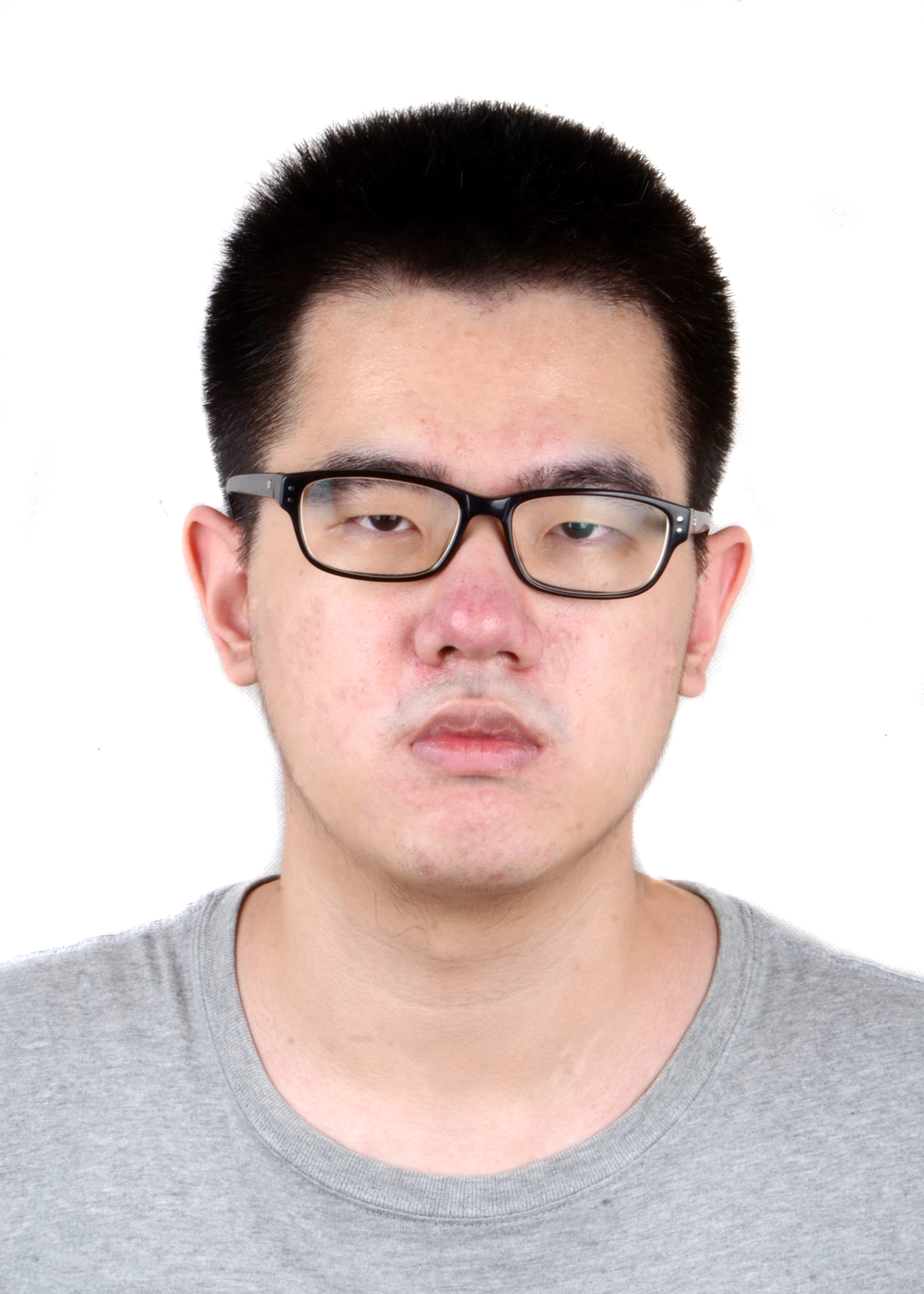}}]{Hanjie Li}
received the B.E. degree in Computer Science and Technology from Beijing
Institute of Technology (BIT) in 2022. He will pursue the master degree in Computer Science and Technology at Beijing Institute of Technology (BIT) from 2022.
His research interests include  graph neural networks and reinforce learning.
\end{IEEEbiography}

\begin{IEEEbiography}[{\includegraphics[width=1in,height=1.25in,clip,keepaspectratio]{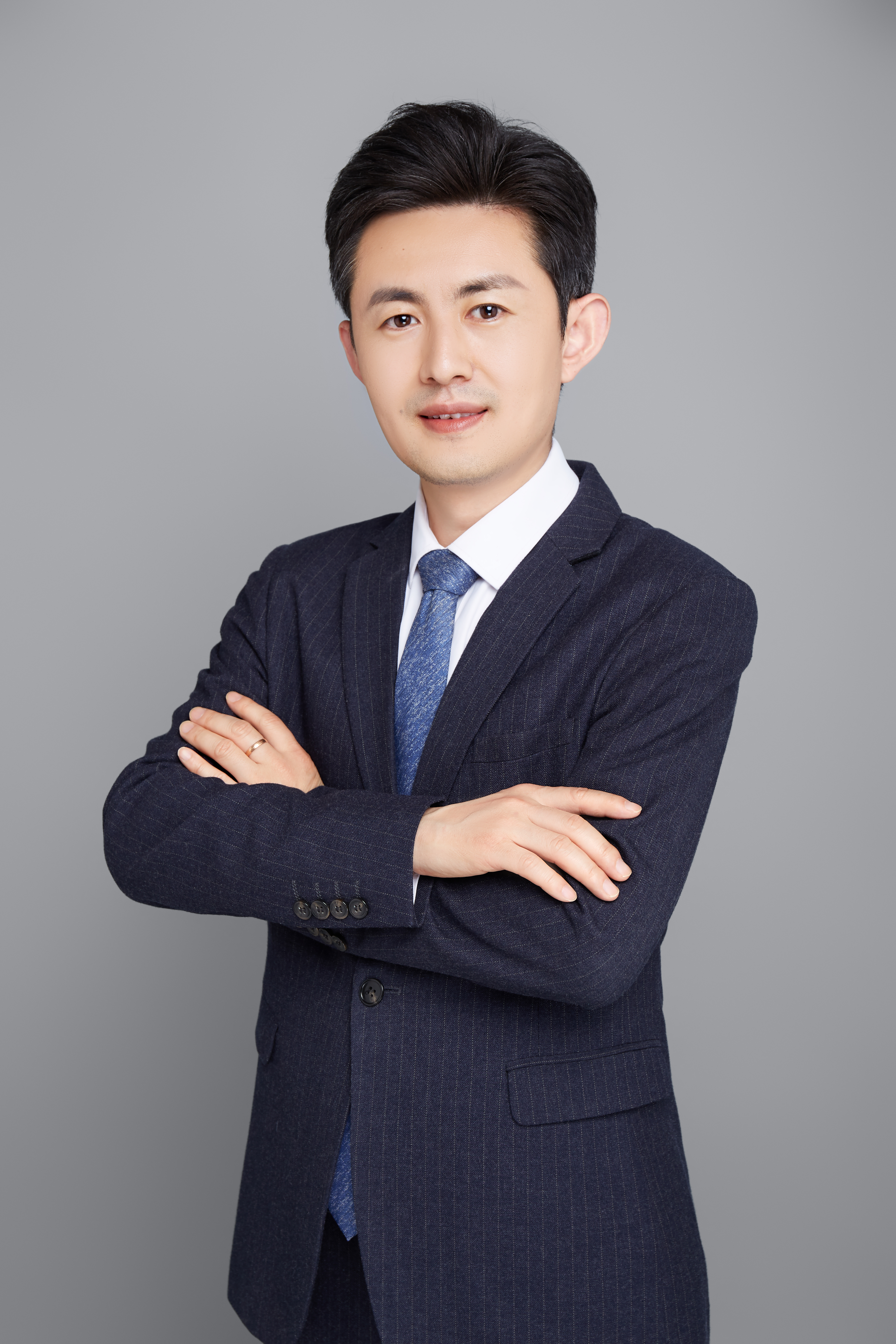}}]{Chengsheng Li}
received the B.E. degree from the University of Electronic Science and Technology of China (UESTC) in 2008 and the Ph.D. degree in pattern recognition and intelligent system from the Institute of Automation, Chinese Academy of Sciences, in 2013. During his Ph.D., he once studied as a Research Assistant with The Hong Kong Polytechnic University from 2009 to 2010. He is currently a Professor with the Beijing Institute of Technology. Before joining the Beijing Institute of Technology, he worked with IBM Research, China, Alibaba Group, and UESTC. He has more than 90 refereed publications in international journals and conferences, including IEEE TRANSACTIONS ON PATTERN ANALYSIS AND MACHINE INTELLIGENCE, IEEE TRANSACTIONS ON IMAGE PROCESSING, IEEE TRANSACTIONS ON NEURAL NETWORKS AND LEARNING SYSTEMS, IEEE TRANSACTIONS ON COMPUTERS, IEEE TRANSACTIONS ON MULTIMEDIA, PR, CVPR, AAAI, IJCAI, CIKM, MM, and ICMR. His research interests include machine learning, data mining, and computer vision. He won the National Science Fund for Excellent Young Scholars in 2021.
\end{IEEEbiography}

\begin{IEEEbiography}[{\includegraphics[width=1in,height=1.25in,clip,keepaspectratio]{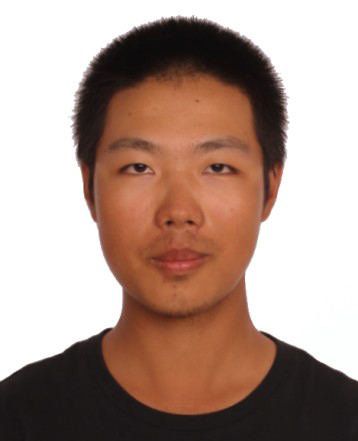}}]{Kaituo Feng}
received the B.E. degree in Computer Science and Technology from Beijing
Institute of Technology (BIT) in 2022. He will pursue the master degree in Computer Science and Technology at Beijing Institute of Technology (BIT) from 2022. His research interests include graph neural networks and knowledge distillation.
\end{IEEEbiography}

\begin{IEEEbiography}[{\includegraphics[width=1in,height=1.25in,clip,keepaspectratio]{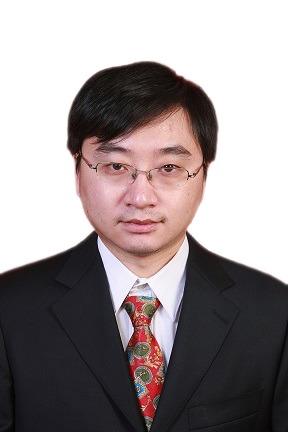}}]{Ye Yuan}
received the B.S., M.S., and Ph.D. degrees in computer science from Northeastern University in 2004, 2007, and 2011, respectively. He is currently a Professor with the Department of Computer Science, Beijing Institute of Technology, China. He has more than 100 refereed publications in international journals and conferences, including VLDBJ, IEEE TRANSACTIONS ON PARALLEL AND DISTRIBUTED SYSTEMS, IEEE TRANSACTIONS ON KNOWLEDGE AND DATA ENGINEERING,SIGMOD, PVLDB, ICDE, IJCAI, WWW, and KDD. His research interests include graph embedding, graph neural networks, and social network analysis. He won the National Science Fund for Excellent Young Scholars in 2016.
\end{IEEEbiography}

\begin{IEEEbiography}[{\includegraphics[width=1in,height=1.25in,clip,keepaspectratio]{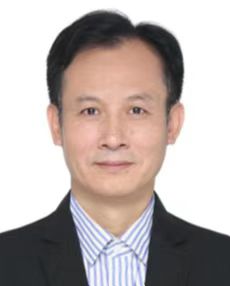}}]{Guoren Wang}
received the B.S., M.S., and Ph.D. degrees in computer science from Northeastern University, Shenyang, in 1988, 1991, and 1996, respectively. He is currently a Professor with the School of Computer Science and Technology, Beijing Institute of Technology, Beijing, where he has been the Dean since 2020. He has more than 300 refereed publications in international journals and conferences, including VLDBJ, IEEE TRANS-ACTIONS ON PARALLEL AND DISTRIBUTED SYSTEMS, IEEE TRANSACTIONS ON KNOWLEDGE AND DATA ENGINEERING, SIGMOD, PVLDB, ICDE, SIGIR, IJCAI, WWW, and KDD. His research interests include data mining, database, machine learning, especially on high-dimensional indexing, parallel database, and machine learning systems. He won the National Science Fund for Distinguished Young Scholars in 2010 and was appointed as the Changjiang Distinguished Professor in 2011.
\end{IEEEbiography}

\begin{IEEEbiography}[{\includegraphics[width=1in,height=1.25in,clip,keepaspectratio]{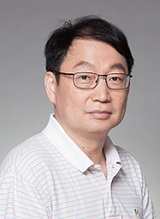}}]{Hongyuan Zha}
received the PhD degree in
scientific computing from Stanford University in
1993. He is a Presidential Chair Professor of the
Chinese University of Hong Kong, Shenzhen. He
has been working on information retrieval, machine
learning applications, and numerical methods.
He received the Leslie Fox Prize (1991, second
prize) of the Institute of Mathematics and its
Applications, the Outstanding Paper Awards of
the 26th International Conference on Advances
in Neural Information Processing Systems (NIPS
2013), and the Best Student Paper Award (advisor) of the 34th ACM SIGIR International Conference on Information Retrieval (SIGIR 2011).
\end{IEEEbiography}

\end{document}

%% file: sections/abstract.tex
\begin{abstract}
Graph structured data often possess dynamic characters in nature, such as the addition of links and nodes, in many real-world applications.  
Recent years have witnessed the increasing attentions paid to dynamic graph neural networks for modelling graph data.
However, almost all existing approaches operate under the assumption that, upon the establishment of a new link, the embeddings of the neighboring nodes should undergo updates to learn temporal dynamics.
Nevertheless, these approaches face the following limitation: If the node introduced by a new connection contains noisy information, propagating its knowledge to other nodes becomes unreliable and may even lead to the collapse of the model.
In this paper, we propose \textbf{Ada-DyGNN}: a robust knowledge \textbf{Ada}ptation framework via reinforcement learning for \textbf{Dy}namic \textbf{G}raph \textbf{N}eural \textbf{N}etworks. 
In contrast to previous approaches, which update the embeddings of the neighbor nodes immediately after adding a new link, Ada-DyGNN adaptively determines which nodes should be updated.
Considering that the decision to update the embedding of one neighbor node can significantly impact other neighbor nodes, we conceptualize the node update selection as a sequence decision problem and employ reinforcement learning to address it effectively. By this means, we can adaptively propagate knowledge to other nodes for learning robust node embedding representations. 
To the best of our knowledge, our approach constitutes the first attempt to explore robust knowledge adaptation via reinforcement learning  specifically tailored  for dynamic graph neural networks. Extensive experiments on three benchmark datasets demonstrate that Ada-DyGNN achieves the state-of-the-art performance. In addition, we conduct experiments by introducing different degrees of noise into the dataset,  quantitatively and qualitatively illustrating the robustness of Ada-DyGNN.  The source code of this work is available at \href{https://github.com/BitLhj/Ada-DyGNN/}{https://github.com/BitLhj/Ada-DyGNN/}
\end{abstract}

%% file: sections/intro.tex
\IEEEraisesectionheading{\section{Introduction}\label{sec:introduction}}
\IEEEPARstart{G}{raph} structured data are ubiquitous in a variety of domains, such as the Internet and the world-wide web \cite{albert1999diameter,park2003hyperlink,pei2019geom}, social networks \cite{sala2010measurement,tang2008arnetminer,xia2021deepis},  scientific citation network \cite{an2021enhancing, feng2022free, van2014citnetexplorer}, bioinformatics \cite{have2013graph, guo2021few, zhang2021graph}, and so on. To better model graph structured data, graph neural networks have recently attracted increasing attention because of their advantages in dealing with complex relations or interactions.
So far, many graph neural network approaches have been proposed in the past decade \cite{henaff2015deep,zhou2020graph,you2019position,li2016gated,zhang2019heterogeneous,wang2019heterogeneous,xinyi2018capsule,liu2020towards}. The representative works include graph attention networks (GAT) \cite{velivckovic2017graph}, GraphSAGE \cite{hamilton2017inductive}, graph convolutional networks (GCN) \cite{kipf2016semi}, etc.

Graph neural networks mentioned above are originally designed for static graphs. However, graph structured data are often dynamic in nature in many real-world applications \cite{liu2019real,yu2017spatio,beladev2020tdgraphembed,lu2019temporal,etemadyrad2021deep,geng2021dynamic,wen2022trend}. Thus, these static graph neural network models often fail in handling such graph data, due to their oversight of the temporal evolution.  To address this issue, several dynamic graph neural networks have been successively proposed to learn the temporal dynamics \cite{trivedi2019dyrep,goyal2020dyngraph2vec,heidari2020evolving,zhou2020data,xu2019adaptive,xue2020modeling,trivedi2017know,rossi2020temporal}.
For instance, DyRep \cite{trivedi2019dyrep} formulates  
representation learning as a latent mediation process.
It proposes a deep temporal point process model with two-time scales to effectively capture the temporal dynamics of observations.
TGN \cite{rossi2020temporal} proposes an efficient framework to combine a memory-related module and graph-based operators for dynamic graphs. 
DyGNN \cite{ma2020streaming} presents an approach for learning node representations when new links are added in the graph. It focuses on modeling the sequential information of edges and the time intervals between interactions to effectively propagate new information to the influenced nodes.

\begin{figure}
    \centering 
    \includegraphics[width=0.99\linewidth]{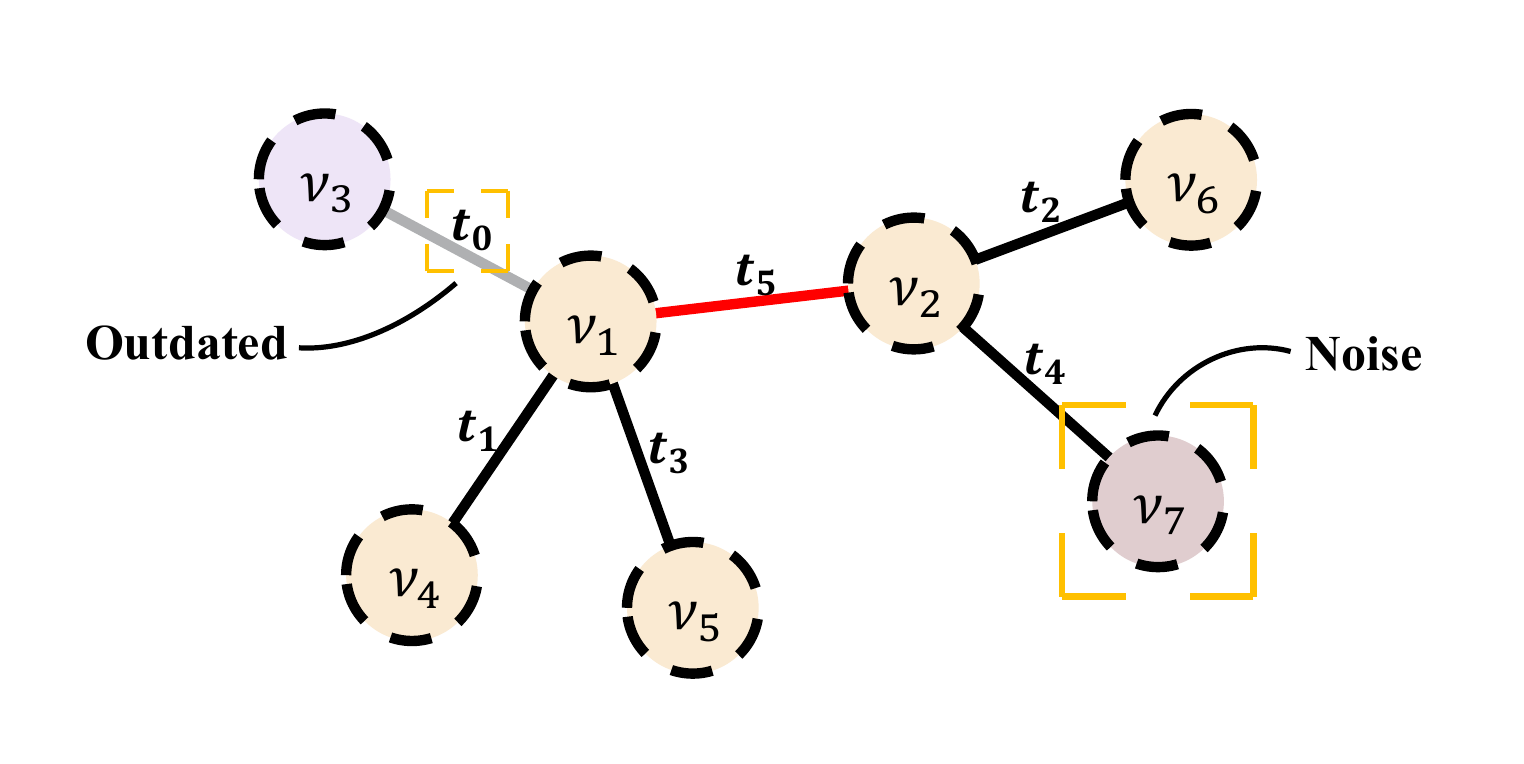}
    \caption{An illustration of noise and outdated links. A new link is built between $v_1$ and $v_2$ at timestamp $t_5$. The link between $v_1$ and $v_3$ is outdated (assume $t_0$ is too much earlier than $t_5$). $v_7$ is the node that contains noisy information. 
    }
    \label{example}
\end{figure}

The aforementioned temporal graph neural network models have achieved promising performance on dynamic graphs of various domains. Typically, these models assume that the embeddings of neighbor nodes need to be updated to capture temporal dynamics once new links are added.
As shown in Fig. \ref{example}, a new edge $t_5$ is established between interactive nodes $v_1$ and $v_2$. Previous methods usually aggregate the knowledge from  the neighbor nodes of the two interactive nodes, 
and then propagate the knowledge to update their embeddings \cite{rossi2020temporal,kumar2019predicting,xu2020inductive}. 
However, in many real-world dynamic graph applications, e.g., recommendation systems, nodes often contain noise, and outdated links between nodes may exist.
In a recommendation system, users and items can be represented as the nodes of a graph. When an user clicks on an item, a link between them is established. 
An item clicked accidentally or mistakenly by an user can be treated as noise. 
Moreover, if a link exists for a long period (e.g., ten years), it is very likely that the link relation between the user and the item is outdated.
In such cases, the aforementioned methods suffer from the following limitations: if a neighbor node contains noisy information (e.g., $v_7$ in Fig. \ref{example}), propagating its knowledge to other nodes based on the existing message-passing mechanism is evidently unreliable and may even lead to the collapse of the model.
If certain existing linkage relations between nodes happen to be out of date, e.g., the edge $t_0$ between $v_1$ and $v_3$ in Fig. \ref{example}, it is no longer suitable to propagate knowledge of $v_3$ and $t_0$ due to the obsolete information, when the new edge $t_5$ is built.
Thus, it is necessary to study how to mitigate the impact of noise or outdated links to effectively propagate new information and update node embeddings in dynamic graphs.

In this paper, we propose \textbf{Ada-DyGNN}:  a reinforcement learning based knowledge \textbf{Ada}ptation framework for \textbf{Dy}namic \textbf{G}raph \textbf{N}eural \textbf{N}etworks, to address the above challenges in a unified framework. 
When new connections between nodes are established, Unlike previous approaches that blindly update the embeddings of related nodes (e.g., neighbors), Ada-DyGNN dynamically and adaptively distinguish which nodes should be influenced and updated. 
Since determining whether one node should be updated will influence the subsequent neighbor nodes, we formulate the node update selection  as a sequence decision problem. 
Moreover, the process of sampling which neighbors to update is discrete, posing a challenge for direct optimization through stochastic gradient descent-based methods \cite{wang2019minimax}. 
In light of these, we attempt to address  this problem via reinforcement learning, which excels in optimizing discrete sampling problems and can capture long-term dependencies and global effects. 
To optimize the policy network, we propose a new reward function to encourage the stability of local structures, defined based on neighbor similarity.
In this way, we can adaptively determine when to propagate knowledge to other neighbor nodes, enabling to learn robust node representations. 

The contributions of this work can be summarized as:
\begin{itemize}
\item We put forward Ada-DyGNN: a robust knowledge adaptation framework to capture temporal evolution for dynamic graph neural networks.   
To the best of our knowledge, our approach constitutes the first attempt to study how to adaptively select the nodes to be updated in dynamic graphs. 
\item We develop a reinforcement learning based method to adaptively distinguish which nodes should be updated, thereby avoiding bringing about the negative impact on the embeddings of the nodes.
Moreover, we devise a new reward function for optimizing the policy network, so as to ensure the stability of the local structure in the evolution of dynamic graphs. 
\item Extensive experiments on three benchmark datasets demonstrate the effectiveness of Ada-DyGNN.
Furthermore, we evaluate Ada-DyGNN on a dataset accompanying with  different levels of noise, which clearly illustrates our model is robust to noise.
\end{itemize}

The rest of this paper is organized as follows. We review the related work in Section \ref{sec:related} and introduce the details of the proposed method in Section \ref{sec:method}. The results of experimental evaluation are reported in Section \ref{sec:experiment}, followed by conclusion and future work in Section \ref{sec:conclusion}.

%% file: sections/related_work.tex
\section{related work}
\label{sec:related}
In this section, we will briefly review the related works to our method, including static graph neural networks, dynamic graph neural networks and reinforcement learning.
\subsection{Static Graph Neural Networks}
{
Static graph neural networks have achieved promising results in learning static graph data. A variety of static graph neural networks have been proposed recently \cite{velivckovic2017graph,wu2019simplifying,chiang2019cluster,chen2018fastgcn,zeng2019graphsaint,bieber2020learning,feng2020graph}. 
For example, GCN \cite{kipf2016semi} is a graph convolutional neural network that proposes to conduct the  convolution operation on graph data. 
GAT \cite{velivckovic2017graph}  takes advantage of the self-attention mechanism to aggregate neighbor information with different weights. 
GraphSAGE \cite{hamilton2017inductive} designs an efficient neighbor sampling mechanism for aggregating information in large-scale graphs.
The work in GIN \cite{xu2018powerful} analyzes that the upper bound of the representation ability of graph neural networks is the Weisfeiler-Lehman isomorphism test \cite{weisfeiler1968reduction}, and builds a GNN model that could reach to this upper bound. 
APPNP \cite{klicpera2018predict} devises an efficient message propagation model on graph data based on the personalized PageRank \cite{page1999pagerank}.
The work in \cite{chen2020simple} effectively alleviates the over-smoothing issue and proposes a deep graph neural network, called GCNII.
However, these static graph neural networks fail to learn the temporal patterns on  dynamic graphs and are shown to be sub-optimal when learning on the  dynamic graphs \cite{xu2020inductive,rossi2020temporal}.
}
\subsection{Dynamic Graph Neural Networks}
Dynamic graph neural networks aim to capture the temporal dynamics for updating the node embeddings, when new connections or links between nodes are established. Based on the properties of dynamic graphs, current dynamic graph neural networks can be roughly divided into two categories \cite{nguyen2018continuous,han2021dynamic,kazemi2020representation,zhu2022learnable}: discrete-time based methods and continuous-time based methods.
The earlier works on dynamic graph mainly focus on the discrete-time method, where the dynamic graphs are regarded as a series of snapshots. The static graph neural network models can be then applied to these snapshots \cite{ibrahim2015link,ahmed2016sampling,ahmed2016efficient,wang2020streaming,pareja2020evolvegcn,chang2020continuous}.
For instance, DynGEM \cite{goyal2018dyngem} uses a deep auto-encoder to incrementally generate stable node embeddings over time on dynamic graphs. DynamicTriad \cite{zhou2018dynamic} aims to capture triad structures in dynamic graphs and derive node representations through a triad closure process.

However, the discrete-time methods cannot capture the fine-grained temporal information of dynamic graphs. To solve this problem, many continuous-time models have been proposed in recent years \cite{trivedi2019dyrep,nguyen2018continuous,rossi2020temporal,ma2020streaming,wang2020streaming}. The continuous-time approaches can observe the dynamic graphs as a sequence of edges (i.e., connections or links) in chronological order.  
The works in \cite{nguyen2018continuous,nguyen2018dynamic} leverage random walks to incorporate continuous time information into walk path selection. 
DyGNN \cite{ma2020streaming} utilizes LSTM-like memory cells to update the interacting nodes and all of their neighborhoods when an interaction occurred. 
\cite{rossi2020temporal} puts forward an inductive learning framework, called TGN, which realized efficient parallel processing on dynamic graphs.
Jodie \cite{kumar2019predicting} proposes a coupled recurrent neural network model for dynamic graph, which can learn the trajectories of users and items.
Most of the continuous-time models focus on designing a message function to aggregate historical or neighborhood information to update the node embeddings. 
However, they attempt to update the embeddings of all the neighbor nodes when a new connection is built, but ignore to distinguish which nodes should be influenced and updated. This leads to a sub-optimal solution, when noisy information or outdated links are involved in the dynamic evolution. 
Thus, we focus on investigating a new mechanism in an effort to selectively update nodes for learning robust node embeddings in dynamic graphs.

\begin{table}
  \caption{Notations and Symbols.}
  \label{Notation}
  \begin{tabular} {cp{170pt}}
      \hline
      Notations & Descriptions\\
      \hline
      $\mathbf{V}$  & set of vertices in the graph\\
      $\mathbf{E}(t)$  & edge set in the dynamic graph at timestamp $t$\\
      $\mathbf{\mathcal{N}}_i(t)$ & set of node $v_i$ and its first-order neighbors at timestamp $t$\\
      $\mathbf{\mathcal{N}}^*_i(t)$ &set of node $v_i$'s first-order neighbors at timestamp $t$\\
      $\mathbf{\mathcal{N}}_{s \cup d}(t)$ &union set of $\mathbf{\mathcal{N}}_s$ and $\mathbf{\mathcal{N}}_d$ at timestamp $t$\\
      $\mathbf{\mathcal{N}}^*_{s \cup d}(t)$ &union set of $\mathbf{\mathcal{N}}^*_s$ and $\mathbf{\mathcal{N}}^*_d$ at timestamp $t$\\
      $\mathbf{x}_i(t)$ & the representation of node $v_i$ at timestamp t\\ 
      $\mathbf{m}(t)$ & the interaction message at timestamp t\\
  \hline
  \end{tabular}
\end{table}

\begin{figure*}
  \centering
  \includegraphics[width=0.96\linewidth]{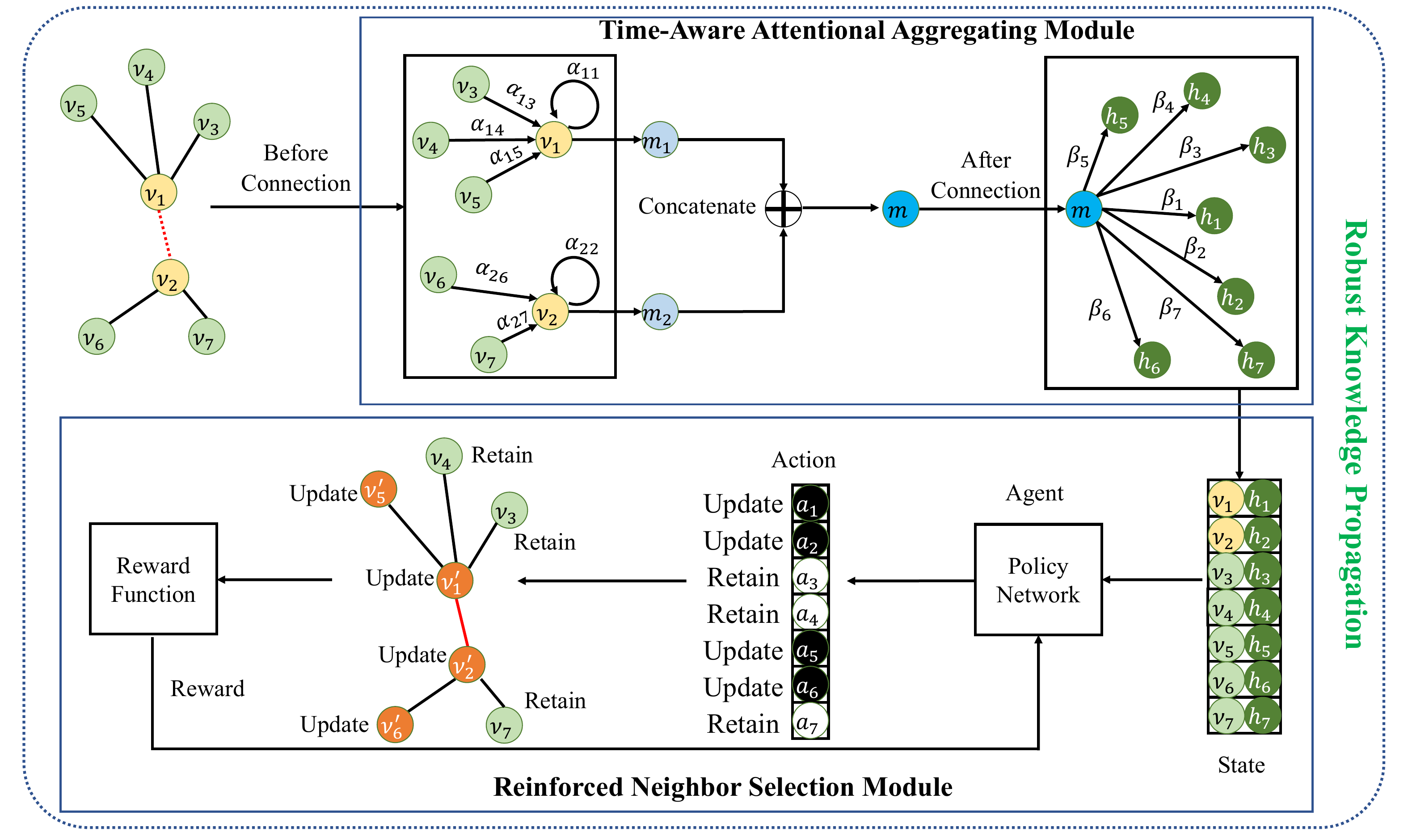}
  \caption{The overall architecture of our Ada-DyGNN method. When a new link between $v_1$ and $v_2$ is added, Ada-DyGNN performs robust knowledge propagation by the following steps: 1) The interaction message $\mathbf{m}_1$ and $\mathbf{m}_2$ are aggregated from $v_1$'s neighborhood $\{v_3, v_4, v_5\}$ and $v_2$'s neighborhood $\{v_6, v_7\}$, respectively; 2) The concatenated interaction message $\mathbf{m}$ propagates the information to $\{v_1,\dots,v_7\}$ to obtain the intermediate states $\{\mathbf{h}_1,\dots,\mathbf{h}_7\}$ based on the time-aware attentional aggregating module, after the connection between $v_1$ and $v_2$ is established; 3) The states are constructed by the node embeddings and the intermediate states, and are sent to the policy network as the inputs; 4) The reinforced neighbor selection module outputs actions to update or retain the embedding of each node; 5) The influenced nodes are updated based on the intermediate embeddings and a MLP. The reward is used to optimize the policy network.}
  \label{overall}
\end{figure*}

\subsection{Reinforcement Learning}
The basic idea of reinforcement learning is to train an agent for decision making by interacting with the environment \cite{kaelbling1996reinforcement,henderson2018deep,angermueller2019model,nachum2018data,nachum2017bridging}. There are mainly two lines of methods  in reinforcement learning \cite{arulkumaran2017deep,li2017deep}: policy-based methods and value-based methods. Value-based methods, such as DQN \cite{fan2020theoretical} and SARSA \cite{zou2019finite}, aim to maximize the expected total reward and take actions according to the expected rewards of actions. Policy-based methods, including REINFORCE \cite{williams1992simple} and Self-Critical Training \cite{rennie2017self}, attempt to train a policy network to generate the policy distributions, from which the actions are sampled.
In addition, Actor-Critic methods \cite{mnih2016asynchronous,haarnoja2018soft,babaeizadeh2016reinforcement,iqbal2019actor} are a hybrid of these two kinds of methods, which makes decisions according to a policy network and estimates the reward by a value function. 
In our method, we attempt to explore reinforcement learning to effectively capture temporal evolution for dynamic graph learning.

%% file: sections/method.tex
\section{Proposed Method}
\label{sec:method}
In this section, we introduce  our Ada-DyGNN in detail. As shown in Fig. \ref{overall}, our model mainly consists of two modules: a time-aware attentional aggregating module aiming to aggregate the neighborhood information; 
a reinforced neighbor selection module intending to adaptively and dynamically determine whether a neighbor node should be updated. We first give some notations and preliminaries before elaborating the details of Ada-DyGNN.

\subsection{Preliminaries}
\noindent{\textbf{Notations}.}  A dynamic graph consists nodes and chronological edges. We define $\mathbf{V}=\{v_1, v_2, \dots, v_n\}$ as the entire node set and $\mathbf{E}(t)=\{(v_{s1},v_{d1},t_1),(v_{s2},v_{d2},t_2),\dots,(v_{sm},v_{dm},t_m),\\ | t_1\leq t_2\leq \dots \leq t_m\leq t\} $ as the sequence of edges until our last observation at time $t$, where $v_{si}$ denotes the source node of the edge built at time $t_i$, and $v_{di}$ denotes the destination node of this edge. $t_i$ denotes the timestamp of interaction between $v_{si}$ and $v_{di}$. $\mathbf{\mathcal{N}}_v(t_i)$ is the set of neighbors of node $v$ at time $t_i$.
The node embeddings at time $t$ are denoted  as   $\mathbf{X}(t)=\{\mathbf{x}_{1}(t),\mathbf{x}_{2}(t),\dots,\mathbf{x}_{n}(t)\}$. {$\mathbf{x}_i(t)\in \mathbb{R}^{d_n}$} denotes  the embedding of node $v_i$ at timestamp $t$, where $d_n$ is the dimension of $\mathbf{x}_i(t)$.
Table \ref{Notation} summarizes the notations used in this paper.
 
\noindent{\textbf{Problem Definition}.}
As the graph evolves, i.e., new edges are continuously added, the inherent properties of nodes will change over time. Dynamic graph neural networks aims to capture the temporal evolution in dynamic graphs to continuously update node embeddings. 
The task can be specifically defined as follows: before a certain timestamp $t$, we  observe a sequence of historical edge sets of a dynamic graph, $\{\mathbf{E}(1),\dots,\mathbf{E}(t)\}$, and temporal node embedding matrices $\{\mathbf{X}(1),\dots,\mathbf{X}(t)\}$. At time $t$, a new edge  $(v_s,v_d,t)$ is added to the graph. 
Note that the new edge may link two existing nodes, or build a connection between a existing node and a new node.
Our goal is to design a robust knowledge propagation mechanism to obtain the  updated embeddings $\mathbf{X}(t)$ of the nodes in the dynamic graph. 
Note that when deleting an edge, robust knowledge adaptation can be resolved similarly to edge addition. Here, we mainly introduce the method when the edge is added. 

\subsection{Time-Aware Attentional Aggregating Module} 
\label{aggregate}
To utilize the temporal information, we construct a time-aware attentional aggregating module to propagate neighborhood information. 
Specifically, when a new interaction $(v_s,v_d,t)$ occurs,  we first calculate the message $\mathbf{m}_s(t)$ and $\mathbf{m}_d(t)$ for nodes $v_s$ and $v_d$, respectively.
 We mainly introduce how to calculate $\mathbf{m}_s(t)$. The method for $\mathbf{m}_d(t)$  is the same.
We set $v_s$ as the central node, and assume the historical interactions on $v_s$ are $(v_s,v_1,t_1)$, $(v_s,v_2,t_2)$, $\dots$, $(v_s,v_k,t_k)$. Note that we do not distinguish the direction of interaction on neighborhoods. The time intervals for the neighbors of the central node are $\Delta t_1=t-t_1, \Delta t_2=t-t_2, \dots, \Delta t_k=t-t_k$. The time interval on the central node is 0. Based on the fact that the impact of interactions with different time intervals can vary to a great extent, a newer interaction should have greater influence on the graph than an older one. Thus we utilize a simple time-decay function $\phi(\Delta t)=\frac{1}{1+\Delta t}$ to reach this goal.
Our time-aware attentional aggregating module combines  the attention mechanism with temporal-related prior information, in order to more effectively aggregate neighborhood information for node embedding learning. 
The time-aware attentional coefficients can be computed as:
\begin{equation}
\label{alpha}
  \alpha_{si}=\frac{\exp (\sigma(\mathbf{a}^{\mathrm{T}}[ \mathbf{W}_g{\mathbf{x}_{s}(t)}\Vert\phi(\Delta t_i) \mathbf{W}_g{\mathbf{x}_{i}(t)}] ))}  {\sum_{j\in \mathbf{\mathcal{N}}_{s}(t)}^{} \exp (\sigma(\mathbf{a}^{\mathrm{T}}[ \mathbf{W}_g{\mathbf{x}_{s}(t)}\Vert\phi(\Delta t_j) \mathbf{W}_g{\mathbf{x}_{j}}(t)] )) } ,
\end{equation}
where $\alpha_{si}$ denotes the time-aware attentional coefficient between node $v_s$ and  $v_i$,
and it measures the importance of $v_i$ to $v_s$ by additionally taking the time interval into consideration.
$\mathbf{W}_g \in \mathbb{R}^{\frac{d_m}{2} \times d_n}$ is a learnable weight matrix.
$\mathbf{a} \in \mathbb{R}^{d_m}$ is a learnable weight vector. 
$\sigma$ is the LeakyReLU activation function.
 $\Vert$ represents the concatenation operation. 

Then, we aggregate the neighborhood information to calculate the message  of source node $v_s$ as:
\begin{equation}
\label{TGAT1}
  \mathbf{m}_s(t) = \sigma_1 (\sum_{i \in \mathbf{\mathcal{N}}_{s}(t)}^{}\alpha_{{s}{i}}\phi(\Delta t_i) \mathbf{W}_g{\mathbf{x}_i(t)} ),
\end{equation}
 where {$\mathbf{m}_s(t)$} is the message of source node $v_s$, with the dimension of $\frac{d_m}{2}$.
 $\sigma_1$ is the ReLU activation function.

Similarly, we calculate the message   of the destination node $v_d$ in the same way as:
\begin{equation}
\label{TGAT2}
  \mathbf{m}_d(t) = \sigma_1 (\sum_{i \in \mathbf{\mathcal{N}}_{d}(t)}^{}\alpha_{{d}{i}}\phi(\Delta t_i) \mathbf{W}_g{\mathbf{x}_i(t)} ),
\end{equation}
where $\mathbf{m}_d(t)$ is the message   of the destination node $v_d$, with the dimension of $\frac{d_m}{2}$.

Next, we integrate the two messages $\mathbf{m}_s(t)$ and $\mathbf{m}_d(t)$, together with the edge feature $\mathbf{e}(t)$ as the interaction message at timestamp $t$.
\begin{equation}
  \mathbf{m}(t) = \mathbf{m}_s(t) || \mathbf{m}_d(t)||\sigma_1(\mathbf{W}_{e}\mathbf{e}(t)),
  \label{mes}
\end{equation}
where $\mathbf{W}_{e}\in \mathbb{R}^{d_m \times d_e}$ is a learnable weight matrix. Note that $e(t)$ is initialized as the zero vector when the edge feature is not available in the dynamic graph.
The interaction message {$\mathbf{m}(t) \in \mathbb{R}^{2d_m}$} encompasses information from the edge between $v_s$ and $v_d$, as well as their respective neighborhoods at timestamp $t$.


Upon establishing the connection between $v_s$ and $v_d$, we compute the intermediate states of both central nodes and neighbor nodes by leveraging the interaction message $\mathbf{m}(t)$. We propose the following time-aware attention mechanism to discern the influence of the interaction message $\mathbf{m}(t)$ on each node:
\begin{equation}\label{beta}
  \beta_{i}=\frac{\exp (\sigma_2(\phi(\Delta t_i)\mathbf{x}_i(t)\mathbf{W}_p\mathbf{m}(t) ))}  {\sum_{j\in \mathbf{\mathcal{N}}_{s \cup d}(t)} \exp (\sigma_2(\phi(\Delta t_j)\mathbf{x}_j(t)\mathbf{W}_p\mathbf{m}(t) ))} ,
\end{equation}

\begin{equation}
\label{propagate}
\mathbf{h}_{i}(t) = \beta_{i}\mathbf{W}_p\mathbf{m}(t),
\end{equation}
where $\beta_{i}$ is the attentional coefficient of node $v_i$. It
 measures the influence of the interaction message $\mathbf{m}(t)$ on $v_i$.
$\sigma_2$ is the Sigmoid activation function.
{$\mathbf{h}_{i}(t) \in \mathbb{R}^{d_n}$} is the intermediate state of  $v_i$. It will be sent to our reinforced neighbor selection module as a part of the inputs.  {$\mathbf{W}_p \in \mathbb{R}^{d_n\times 2 d_m}$} is a learnable weight matrix. Note that we use the term $\mathbf{x}_i(t)\mathbf{W}_p\mathbf{m}(t)$ to calculate $\beta_i$ in Eq. (\ref{beta}). The reason for choosing the inner product function is that $\mathbf{x}_i(t)$ and $\mathbf{m}(t)$ belong to different spaces, as indicated by Eq. (\ref{TGAT1}) and (\ref{TGAT2}). In Eq. (\ref{TGAT1}) and (\ref{TGAT2}), we first learn a matrix $\mathbf{W}_g$ to map $\mathbf{x}_i(t)$ to another feature space, and then aggregate neighbor information to obtain the message $\mathbf{m}(t)$. 
Therefore, in order to effectively calculate $\beta_i$, we aim to learn another transformation matrix $\mathbf{W}_p$ to project $\mathbf{m}(t)$ to the same space as $\mathbf{x}_i(t)$.
Subsequently, we utilize the inner product between $\mathbf{x}_i(t)$ and $\mathbf{W}_p\mathbf{m}(t)$ to measure their similarity.

\subsection{Reinforced Neighbor Selection Module} 
In a dynamic graph, as new edges are added continuously, the inherent properties of nodes will vary over time.
Thus, the embeddings of the two interactive nodes might need to be updated, and the neighbors of central nodes (i.e., the two interactive nodes) might be influenced. 
Previous studies, such as \cite{ma2020streaming}, attempt to update the embeddings of the central nodes and all of their neighborhoods once a new edge is established. 
However, such a learning strategy is not reasonable  in many real-world applications, due to the following reasons: 
if the neighbor node contains noisy information, it might not be helpful to propagate its information to other nodes. In contrast, such a propagation mechanism could lead to the collapse of the learning model.
Moreover,  some linkage relations between nodes might happen to be out of date as the graph is evolving, thus it might be not suitable to propagate new information between them.

Based on the above consideration, we attempt to adaptively select neighbors to update.
Note that, since sampling which of neighbors to update is discrete, we could not optimize it through stochastic gradient descent based methods \cite{wang2019minimax,chen2021user}. More importantly, the process of deciding whether neighbor nodes should be updated or retained can be regraded as a sequence decision problem.
Thus, we 
address this problem via reinforcement learning, which is good at optimizing discrete problem and can capture long-term dependencies and global effects for making decisions. As shown in Fig. \ref{overall}, we construct the environment
by the dynamic graph and the time-aware attentional aggregating module. 
When a new interaction occurs, the agent receives states from the environment,  which are the concatenation of the interaction messages and the intermediate embeddings. 
Then, the agent takes actions based on current states and a learned policy network, which can determine whether to update or retain the embedding for each node.
After that, new embeddings of the influenced nodes can be obtained based on the intermediate state and a MLP.
Finally, we maximize the reward to optimize the policy network.  
Reinforcement learning mainly consists of three elements: state, action, reward.
Next, we will introduce them in detail.  

\noindent{\textbf{State.}}
When an interaction $(v_s,v_d,t)$ occurs at time $t$, we calculate the state $\mathbf{s}_i(t)$ for each node $v_i \in \mathbf{\mathcal{N}}_{s \cup d}(t)$. The state $\mathbf{s}_i(t)$ is composed of the following two kinds of features:
\begin{itemize}
    \item {the node embedding $\mathbf{x}_i(t)$.}
    \item the intermediate state $\mathbf{h}_i(t)$ of node $v_i$.
\end{itemize}
The former summarizes the history information of node $v_i$ until timestamp $t$.
The latter represents the received message with distinguishing impact.
The state $\mathbf{s}_{i}(t)$ can be expressed as:
\begin{equation}
    \mathbf{s}_{i}(t)=\mathbf{h}_{i}(t)||\mathbf{x}_i(t).
    \label{state}
\end{equation}

\noindent{\textbf{Action.}}
 The action of the agent is denoted as $a_i \in \{0,1\}$ , representing whether the agent decides to update the node $v_i$ or not. $a_i=1$ stands for the agent decides to update the representation of node $v_i$, while $a_i=0$ means that the agent decides to keep the representation of node $v_i$. $a_i$ is sampled according to a probability distribution produced by a learned policy network $\pi$, which consists of two fully-connected layers. Formally, the policy $\pi(\mathbf{s}_i(t))$ is calculated as follows:
\begin{equation}
    \pi(\mathbf{s}_{i}(t)) = \sigma_2(\mathbf{W}_{1}\sigma_1(\mathbf{W}_{2}\mathbf{s}_{i}(t)))),
    \label{policy}
\end{equation}
where $\sigma_1$ and $\sigma_2$ are ReLU and sigmoid activation functions respectively.
{ $\mathbf{W}_{1}\in \mathbb{R}^{1\times d_n}$ and $\mathbf{W}_{2} \in \mathbb{R}^{d_n\times 2d_n}$ }are two learnable weight matrices.
When a node $v_i \in \mathbf{\mathcal{N}}_{s \cup d}$ is determined to be updated, we utilize  its previous embedding $\mathbf{x}_{i}(t)$ and its intermediate state $\mathbf{h}_{i}(t)$ to obtain its updated embedding. The new embedding $\mathbf{x}_{i}(t+)$ of $v_i$ can be calculated as:
\begin{equation}
\label{update_neighbor}
  \mathbf{x}_{i}(t+) = \sigma_1(\mathbf{W}_{u}(\mathbf{x}_{i}(t)||\mathbf{h}_{i}(t)),
\end{equation} 
where {$\mathbf{W}_{u}\in \mathbb{R}^{d_n\times 2d_n}$} is a learnable weight matrix.
If the agent decides to retain its embedding, its embedding $\mathbf{x}_{i}(t+)$ will be kept as:
\begin{equation}
    \mathbf{x}_{i}(t+)=\mathbf{x}_{i}(t).
\end{equation}

\noindent Note that the initial embedding of $\mathbf{x}_i$, i.e., $\mathbf{x}_i(0)$, is initialized randomly.

\noindent{\textbf{Reward.}}
Given that graph datasets typically include topological relationships, we leverage the inherent structure of the graph to define rewards for the generalization of our method.
Motivated by the previous work \cite{goyal2018dyngem} that has proven the importance of the stability of node embeddings for dynamic graphs and defined the stability by directly measuring the embedding difference between adjacent snapshots, we believe that a high similarity between node embeddings can indicate better retention of historical topological information.
Thus, we define the stability of the local structure by requiring the embeddings of the center node and its neighbor nodes to be as similar as possible. 
Meanwhile, we regard the stability of the local structure as our reward, defined as: 
\begin{equation}
  r = \frac{\sum\limits_{i\in\mathbf{\mathcal{N}}^*_{s}(t)}^{}\!\!cos(\mathbf{x}_{s}(t+),\mathbf{x}_{i}(t+))}{|\mathbf{\mathcal{N}}^*_{s}(t)|} + \frac{\sum\limits_{i\in\mathbf{\mathcal{N}}^*_{d}(t)}^{}\!\!cos(\mathbf{x}_{d}(t+),\mathbf{x}_{i}(t+))}{| \mathbf{\mathcal{N}}^*_{d}(t)|},
  \label{rewardfunc}  
\end{equation} 
where  $\mathbf{\mathcal{N}}^*_s(t)$ and $\mathbf{\mathcal{N}}^*_d(t)$ represent the neighbor set of node $v_s$ and node $v_t$ at timestamp $t$, respectively. $|\mathbf{\mathcal{N}}^*_{s}(t)|$ and $| \mathbf{\mathcal{N}}^*_{d}(t)|$ represent the size of  $\mathbf{\mathcal{N}}^*_{s}(t)$ and $\mathbf{\mathcal{N}}^*_{d}(t)$, respectively. $cos(\cdot)$ is the cosine similarity function. 
The first part of the reward measures the similarity between the embeddings of source node $v_s$ and its temporal neighbor nodes at timestamp $t$, while the second part measures the similarity between the embeddings of destination node $v_d$ and its temporal neighbor nodes at timestamp $t$. 
By maximizing the reward $r$, we can preserve the  historical topological information and maintain the stability of local structures, so as to achieve robust knowledge adaptation in dynamic graphs.

\begin{figure}
    \centering 
    \includegraphics[width=0.99\linewidth]{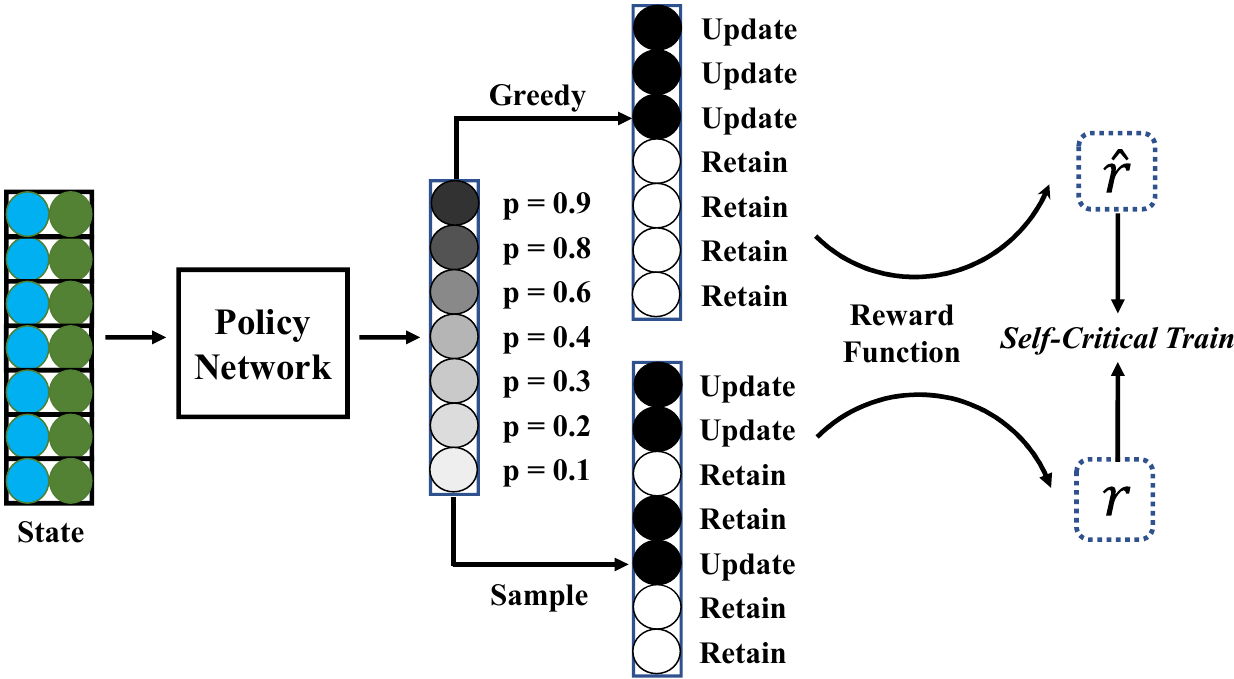}
    \caption{The training of our policy network. }
    \label{SCST}
\end{figure}

\noindent{\textbf{Optimization.}}
We optimize our policy network by self-critical training \cite{rennie2017self},  
which has been widely used on sequential tasks such as sequential recommendation \cite{chen2021user}. We adopted the self-critical training because it adds an inference time baseline on the basis of the REINFORCE algorithm \cite{williams1992simple} to normalize the rewards, which could speed up convergence.
Fig. \ref{SCST} shows the training details of our policy network.
We take actions independently with two different strategies as shown in Fig. \ref{SCST}.
The first strategy is a baseline, which greedily update node with the probability great than or equal to 0.5. The second strategy samples the actions from the probability distributions, which is the actual action we taken on propagation. The acquired rewards for the greedy strategy and sampling strategy are $\hat{r}$ and $r$, respectively. 
Finally, we optimize the policy network by:
\begin{equation}
\label{rl_optimize}
    \theta \leftarrow \theta + \eta\frac{1}{\left| \mathbf{\mathcal{N}}_{s \cup d}(t) \right|} \sum_{i\in\mathbf{\mathcal{N}}_{s \cup d}(t)}^{}(r-\hat{r})\nabla_\theta\log\pi_\theta(\mathbf{s}_i(t)),
\end{equation}
where $\theta$ are the learned parameters of the policy network $\pi$, $\left| \mathbf{\mathcal{N}}_{s \cup d} \right|$ is the size of $\mathbf{\mathcal{N}}_{s \cup d}(t)$, and $\eta$ is the leaning rate.

\subsection{Model Training}
The time-aware attentional aggregating module and the reinforced neighbor selecting module are jointly optimized  during training. 
Algorithm \ref{algorithm} lists the details of the training procedure of our method. 
At the beginning of each epoch, the node embeddings are randomly initialized since the raw node features are unavailable. Following \cite{rossi2020temporal,xu2020inductive}, we randomly initialize the embeddings with a fixed seed to ensure the initialized values are the same at timestampt 0. It is important to note that the node embeddings are not treated as learnable parameters. Instead, they are initialized with a fixed value at timestamp 0, and subsequently dynamically updated by our Ada-DyGNN as the dynamic graph evolves.
Especially, when a new edge $(v_s,v_d,t)$ is established, we first calculate the interaction message by the time-aware attentional aggregating module.  
After that, for each node in $\mathbf{\mathcal{N}}_{s \cup d}(t)$, we calculate its intermediate state. 
Then, we concatenate the node embedding and the intermediate state to obtain the RL state. The state is then fed into the policy network to generate the policy distributions.
According to the policy distributions, we calculate the rewards to optimize the policy network with the self-critical training method. Finally, we sample a negative node $v_n$, and construct a negative edge $(v_s,v_n,t)$. We use the cross-entropy loss to optimize  the time-aware attentional aggregating module, defined as:
\begin{align}\nonumber
  L_{ce}&=-log(\sigma_2(\mathbf{x}_{s}(t+)^\mathrm{T}\mathbf{x}_{d}(t+)))\\
  &\ \ \ \  - log(\sigma_2(1-\mathbf{x}_{s}(t+)^\mathrm{T}\mathbf{x}_{n}(t+))),
  \label{lossCE}
\end{align}
where $\sigma_2$ is the sigmoid activation function.

Note that the selection of node updates in the RL component is restricted to the neighbors of the central node; that is, we do not choose the negative sample for updating when new links are added.
Consequently, we do not incorporate negative examples when calculating the reward function in Eq. (\ref{rewardfunc}).

\begin{algorithm}[t]
  \caption{The Proposed Ada-DyGNN}  
  \label{algorithm}
  \begin{algorithmic}  
    \STATE 
    \STATE {\textbf{Require}: The entire chronological edge list $\mathbf{E}$, the node set $\mathbf{V}$}.
    \STATE {\textbf{For} each Epoch
    \textbf{do} 
    \STATE \hspace{0.5cm}{Initialize the node embeddings with a fixed seed.}}
    \STATE \hspace{0.5cm}{\textbf{For} $(v_s,v_d,t) \in \mathbf{E}$} \textbf{do} 
    \STATE \hspace{1.0cm}Calculate the interaction message by (\ref{mes}).
    \STATE \hspace{1.0cm}\textbf{For} {$v_i \in \mathbf{\mathcal{N}}_{s \cup d}(t)$} \textbf{do} 
    \STATE \hspace{1.5cm}Calculate intermediate state $\mathbf{h}_{i}(t)$ by  (\ref{propagate}).
    \STATE \hspace{1.5cm}Calculate state $\mathbf{s}_i(t)$ by (\ref{state}).
    \STATE \hspace{1.5cm}Generate policy $\pi(\mathbf{s}_i(t))$\ by (\ref{policy}).
    \STATE \hspace{1.0cm}\textbf{End For}
    \STATE \hspace{1.0cm}Calculate baseline reward $\hat{r}$ with a greedy strategy by (\ref{rewardfunc}).
    \STATE \hspace{1.0cm}Take actions with sampling strategy and acquire reward $r$ by (\ref{rewardfunc}).
    \STATE \hspace{1.0cm}Optimize the policy network by (\ref{rl_optimize}).
    \STATE \hspace{1.0cm}Sample a negative node $v_n$ from $\mathbf{V}$.
    \STATE \hspace{1.0cm}Optimize the parameters of aggregating module and $\mathbf{W}_{u}$ in (\ref{update_neighbor}) by (\ref{lossCE}).
    \STATE \hspace{0.5cm}\textbf{End For}    
    \STATE \textbf{End For} 
  \end{algorithmic}  
\end{algorithm}  

%% file: sections/experiment.tex
\section{Experiments}
\label{sec:experiment}
To demonstrate the effectiveness of our model, we conduct extensive experiments on three real-world temporal graph datasets for future link prediction tasks.
To further evaluate our model, we compare it with the state-of-the-art baselines.

\begin{table}[b]
\centering
  \caption{Statistics of three real-world graph datasets.}
  \label{Statistics}
  \begin{tabular}{lccc}
      \toprule
      &UCI & Wikipedia & Reddit\\
      \midrule
      Number of nodes&1899&9227&10984\\
      Number of edges&59835&157474&672447\\
      Time duration  &194 days&1 month & 1 month\\
  \bottomrule
  \end{tabular}
\end{table}

\subsection{Experiment Setup}
\subsubsection{Datasets}
We use three real-world datasets in our experiments: UCI \cite{kunegis2013konect}, Wikipedia \cite{kumar2019predicting} and Reddit \cite{kumar2019predicting}.
The statistics of these three datasets are shown in Table \ref{Statistics}.

{
\textbf{UCI datastet.}
UCI is a directed graph dataset that describes the online message communications among student users on an online platform at the University of California, Irvine.
In the UCI dataset, the nodes represent users, and the edges signify message communications between users on the online platform.
}

{
\textbf{Wikipedia dataset.}
The Wikipedia dataset is a bipartite graph that illustrates editing interactions between users and web pages in a free online encyclopedia Wikipedia.
The nodes represent both users and web pages in Wikipedia.
An interaction in this dataset denotes a user editing a Wikipedia web page.
}

{
\textbf{Reddit dataset.}
The Reddit dataset is also a bipartite graph that records posting interactions in a large online platform of topic communities, Reddit.
The nodes represent Reddit users and sub-reddits. 
When a user posts a passage to a sub-reddit, a new edge is created between the user and the sub-reddit.
}

\subsubsection{Baselines}
We compare our method with 9 baselines: GCN \cite{kipf2016semi}, GraphSAGE \cite{hamilton2017inductive}, GAT \cite{velivckovic2017graph} are three static GNN models, which could not leverage temporal information. For performing the three baselines, we regard the dynamic graphs as a series of static graphs by disregarding temporal information. DyGNN \cite{ma2020streaming}, DyRep \cite{trivedi2019dyrep}, CTDNE \cite{nguyen2018continuous}, TGAT \cite{xu2020inductive}, Jodie \cite{kumar2019predicting},  TGN \cite{rossi2020temporal} are six dynamic graph neural networks. We provide a brief introduction to these methods as follows:
\begin{itemize}
  \item GCN \cite{kipf2016semi} is a neural network that operates the convolution operation on graph structured data. It has made breakthroughs in the semi-supervised node classification task.
  \item GraphSAGE \cite{hamilton2017inductive}  can learn node representations in a large graph with limited computing resources by designing an efficient neighbor sampling mechanism.
  \item GAT \cite{velivckovic2017graph} applies the self-attention mechanism to assign different weights to different neighbors for effectively aggregating neighbor information.
  \item DyGNN \cite{ma2020streaming} is composed of an update component and a propagation component based on memory cells to process dynamic graph data. It could capture the temporal information as the dynamic graph evolves.
  \item DyRep \cite{trivedi2019dyrep} is a representation learning method based on temporal point process for dynamic graphs.
  \item CTDNE \cite{nguyen2018continuous} is a transductive method  for dynamic graphs, which takes advantage of the random walk to obtain the embeddings of the nodes.
  \item TGAT \cite{xu2020inductive} is a  variant of GAT tailored for dynamic graphs. It can infer node representations in the inductive setting.
  \item Jodie \cite{kumar2019predicting} is a coupled recurrent neural network model. It learns embedding from node's trajectory to predict future interactions. 
  \item TGN \cite{rossi2020temporal} is a generic continuous-time inductive framework for temporal graph networks with well-designed  memory modules and graph-based operators.
  \item Ada-DyGNN is our proposed method that {aims to devise a robust knowledge adaptation mechanism for dynamic graphs. Ada-DyGNN can   adaptively select the neighbor nodes to be updated via reinforcement learning, as the graph evolves.
  }
\end{itemize}

\subsubsection{Evaluation Metrics}
Following \cite{ma2020streaming,rossi2020temporal}, 
{
 we evaluate the performance of the compared methods on future link prediction tasks from the following two perspectives:
\begin{itemize}
\item Mean Reciprocal Rank (MRR): We utilize a ranking metric, Mean Reciprocal Rank (MRR), to evaluate the performance. The calculation method can be expressed as follows: for each edge $(v_s,v_d,t)$ in the testing set as the positive sample, we fix $v_s$ and replace $v_d$ by other nodes $v_n$ in the entire node set $V$, as negative samples $(v_s,v_n,t)$. 
Then we rank the $|V|$ samples (one positive sample and $|V|-1$ negative samples) in the descending order by the cosine similarity between node pairs. 
Mathematically, MRR can be expressed as:
\begin{equation}
MRR = \frac{1}{M}\sum_{i=1}^{M}\frac{1}{rank_i}
\end{equation}
where $M$ is the number of edges in the testing set, and $rank_i$ is the rank of the $i^{th}$ testing edge out of all the $|V|$ samples. 
A higher MRR score corresponds to better performance.
\item AP and AUC: We use two popular classification metrics, Average Precision (AP) and Area Under Curve (AUC), to evaluate the performance.
In this context, the link prediction between two nodes is regarded as a binary classification task. Unlike ranking, we generate one negative sample for each testing edge when calculating AP and AUC.
\end{itemize}
Note that AP and AUC measure the positive sample with a single negative sample, while MRR compares it with a mass of negative samples (equal to the size of $V$). 
Therefore, MRR is more challenging than AP and AUC. 
}
\subsubsection{Implementation Details}
We use the Adam optimizer \cite{zhang2018improved} to train our model. 
The early stopping strategy is adopted with a patience on validation sets of 10 epochs. The dropout rate is set to 0.5 and the learning rate is set to 0.0001. 
Both dimensions, $d_n$ and $d_m$, are consistently set to 64 throughout the experiment. 
On the three real-world datasets, the degree of nodes varies dramatically. 
For the sake of the efficiency and parallel processing ability, our reinforced neighbor selection module only samples the most $k=200$ recent interaction neighbors. We also perform the experiments to show the performances of our method with different values of $k$. 
In our experiment, we study our method in both transductive and inductive settings. 
In both settings, we use the first 80\% of the edges as the training set, 10\% of the edges as the validation set and the rest 10\% edges as the testing set. 
Differently, in the inductive setting we predict future edges of nodes never seen in the training set, while in the transductive setting we predict future edges of nodes observed in the training set. 
We run each experiment 10 times, and report the average results.

\begin{table*}
\centering
  \caption{Mean Reciprocal Rank (MRR) of different methods on the three datasets in the transductive and inductive settings for the future link prediction task. '-' does not support the inductive setting. Bold represents the best result.}
  \label{result_mrr}
  \begin{tabular}{l|cc|cc|cc}
      \hline
      &\multicolumn{2}{c|}{UCI}&\multicolumn{2}{c|}{Wikipedia}&\multicolumn{2}{c}{Reddit} \\
      \cline{2-7}
      &Transductive&Inductive&Transductive&Inductive&Transductive&Inductive\\
      \hline
      GCN       &0.0075 $\pm$ 0.0012 &-&0.0068 $\pm$ 0.0004 &-&0.0007 $\pm$ 0.0002&-\\
      GraghSAGE &0.0051 $\pm$ 0.0005 &0.0036 $\pm$ 0.0009&0.0094 $\pm$ 0.0014 &0.0042 $\pm$ 0.0005&0.0022 $\pm$ 0.0006&0.0017 $\pm$ 0.0003\\
      GAT       &0.0087 $\pm$ 0.0013 &0.0074 $\pm$ 0.0005&0.0187 $\pm$ 0.0024 &0.0101 $\pm$ 0.0017&0.0022 $\pm$ 0.0003&0.0018 $\pm$ 0.0005\\\hline
      CTDNE     &0.0081 $\pm$ 0.0027 &-&0.0041 $\pm$ 0.0003 &-&0.0022 $\pm$ 0.0004&-\\
      DyGNN     &0.0108 $\pm$ 0.0001 &-&0.0165 $\pm$ 0.0002 &-&0.0156 $\pm$ 0.0001&-\\
      DyRep     &0.0312 $\pm$ 0.0001 &0.0165 $\pm$ 0.0001&0.3253 $\pm$ 0.0043 &0.3165 $\pm$ 0.0001&0.0378 $\pm$ 0.0001&0.0816 $\pm$ 0.0001\\
      TGAT      &0.0094 $\pm$ 0.0011 &0.0045 $\pm$ 0.0002&0.0558 $\pm$ 0.0045 &0.0118 $\pm$ 0.0015&0.0538 $\pm$ 0.0034&0.0350 $\pm$ 0.0133\\
      Jodie     &0.0116 $\pm$ 0.0004 &0.0086 $\pm$ 0.0018&0.1199 $\pm$ 0.0050 &0.1399 $\pm$ 0.0054&0.0315 $\pm$ 0.0012&0.0413 $\pm$ 0.0049\\
      TGN       &0.0159 $\pm$ 0.0042 &0.0143 $\pm$ 0.0061&0.1829 $\pm$ 0.0036 &0.1786 $\pm$ 0.0035&0.0210 $\pm$ 0.0029&0.0203 $\pm$ 0.0018\\\hline
            Ada-DyGNN &\textbf{0.0989} $\pm$ 0.0073 &\textbf{0.0886} $\pm$ 0.0084&\textbf{0.3323} $\pm$ 0.0053&\textbf{0.3247} $\pm$ 0.0070&\textbf{0.0568} $\pm$ 0.0004& \textbf{0.0981} $\pm$ 0.0014\\
  \hline
  \end{tabular}
\end{table*}

\begin{table*}
\centering
  \caption{Average Precision (AP)  of different methods on the three datasets in the transductive and inductive settings for the future link prediction task. Bold represents the best result.}
  \label{result_ap}
  \begin{tabular}{l|cc|cc|cc}
      \hline
      &\multicolumn{2}{c|}{UCI}&\multicolumn{2}{c|}{Wikipedia}&\multicolumn{2}{c}{Reddit} \\\cline{2-7}
      &Transductive&Inductive&Transductive&Inductive&Transductive&Inductive\\
      \hline
      GCN       &0.5550 $\pm$ 0.0246&-&0.5860 $\pm$ 0.0048&-&0.7650 $\pm$ 0.0049&-\\
      GraphSage &0.5500 $\pm$ 0.0220&0.4763 $\pm$ 0.0119&0.8529 $\pm$ 0.0034&0.8002 $\pm$ 0.0067&0.8241 $\pm$ 0.0220&0.8055 $\pm$ 0.0195\\
      GAT       &0.5645 $\pm$ 0.0021&0.5370 $\pm$ 0.0269&0.9098 $\pm$ 0.0049&0.8619 $\pm$ 0.0057&0.9139 $\pm$ 0.0098&0.9056 $\pm$ 0.0131\\\hline
      CTDNE     &0.4907 $\pm$ 0.0004&-&0.6345 $\pm$ 0.0003&-&0.5115 $\pm$ 0.0002&-\\
      DyGNN     &0.7519 $\pm$ 0.0035&-&0.7912 $\pm$ 0.0002&-&0.8065 $\pm$ 0.0009&-\\
      DyRep     &0.5507 $\pm$ 0.0803&0.5074 $\pm$ 0.0662&0.9480 $\pm$ 0.0013&0.9216 $\pm$ 0.0010&0.9803 $\pm$ 0.0002&0.9568 $\pm$ 0.0026\\
      TGAT     &0.8534 $\pm$ 0.0118&0.7078 $\pm$ 0.0194&0.9472 $\pm$ 0.0030&0.9318 $\pm$ 0.0013&0.9703 $\pm$ 0.0017&0.9471 $\pm$ 0.0051\\
      Jodie      &0.7793 $\pm$ 0.0035&0.7036 $\pm$ 0.0030&0.9519 $\pm$ 0.0009&0.9399 $\pm$ 0.0005&0.9824 $\pm$ 0.0004&0.9668 $\pm$ 0.0007\\
      TGN       &0.8499 $\pm$ 0.0468&0.8140 $\pm$ 0.0217&0.9847 $\pm$ 0.0006&0.9778 $\pm$ 0.0013&0.9871 $\pm$ 0.0002&0.9758 $\pm$ 0.0009\\\hline
            Ada-DyGNN &\textbf{0.9191} $\pm$ 0.0140&\textbf{0.8258} $\pm$ 0.0218&\textbf{0.9898} $\pm$ 0.0004&\textbf{0.9860} $\pm$ 0.0009&\textbf{0.9936} $\pm$ 0.0016&\textbf{0.9880} $\pm$ 0.0029\\
  \hline
  \end{tabular}
\end{table*}

\begin{table*}
\centering
  \caption{Area Under Curve (AUC) of different methods on the three datasets in the transductive and inductive settings for the future link prediction task. Bold represents the best result.}
  \label{result_auc}
  \begin{tabular}{l|cc|cc|cc}
      \hline
      &\multicolumn{2}{c|}{UCI}&\multicolumn{2}{c|}{Wikipedia}&\multicolumn{2}{c}{Reddit} \\\cline{2-7}
      &Transductive&Inductive&Transductive&Inductive&Transductive&Inductive\\
      \hline

      GCN       &0.5620 $\pm$ 0.0210&-&0.5880 $\pm$ 0.0020&-&0.7821 $\pm$ 0.0086&-\\
      GraphSage &0.5666 $\pm$ 0.0134&0.4694 $\pm$ 0.0217&0.8091 $\pm$ 0.0033&0.7511 $\pm$ 0.0036&0.8484 $\pm$ 0.0229&0.8367 $\pm$ 0.0236\\
      GAT       &0.5589 $\pm$ 0.0244&0.5151 $\pm$ 0.0052&0.8840 $\pm$ 0.0066&0.8266 $\pm$ 0.0060&0.9269 $\pm$ 0.0094&0.9220 $\pm$ 0.0128\\\hline
      CTDNE     &0.4738 $\pm$ 0.0005&-&0.5788 $\pm$ 0.0010&-&0.4144 $\pm$ 0.0003&-\\
      DyGNN     &0.7770 $\pm$ 0.0020&-&0.8145 $\pm$ 0.0003&-&0.8332 $\pm$ 0.0002&-\\
      DyRep     &0.5720 $\pm$ 0.0791&0.4879 $\pm$ 0.0794&0.9442 $\pm$ 0.0010&0.9151 $\pm$ 0.0037&0.9798 $\pm$ 0.0003&0.9565 $\pm$ 0.0021\\
      TGAT     &0.8852 $\pm$ 0.0055&0.7386 $\pm$ 0.0129&0.9496 $\pm$ 0.0019&0.9280 $\pm$ 0.0009&0.9755 $\pm$ 0.0016&0.9533 $\pm$ 0.0043\\
      Jodie      &0.7987 $\pm$ 0.0031&0.7044 $\pm$ 0.0018&0.9489 $\pm$ 0.0007&0.9355 $\pm$ 0.0010&0.9819 $\pm$ 0.0008&0.9647 $\pm$ 0.0010\\
      TGN       &0.8478 $\pm$ 0.0459&0.7924 $\pm$ 0.0225&0.9839 $\pm$ 0.0006&0.9766 $\pm$ 0.0013&0.9867 $\pm$ 0.0003&0.9748 $\pm$ 0.0011\\\hline
            Ada-DyGNN &\textbf{0.9259} $\pm$ 0.0112&\textbf{0.8204} $\pm$ 0.0263&\textbf{0.9880} $\pm$ 0.0006&\textbf{0.9837} $\pm$ 0.0013&\textbf{0.9930} $\pm$ 0.0019&\textbf{0.9868} $\pm$ 0.0035\\
  \hline
  \end{tabular}
\end{table*}

\subsection{General Performance }
Table \ref{result_mrr}, \ref{result_ap} and \ref{result_auc} list the experimental results.
As shown in Table \ref{result_mrr},  our Ada-DyGNN model consistently outperforms all the static and dynamic approaches in terms of MRR. This illustrates that designing robust knowledge adaptation mechanism is beneficial to dynamic graph neural networks.
For the transductive setting, our Ada-DyGNN relatively improves the performance  by 217.0\%, 2.2\%, 50.2\% in terms of MRR over DyRep which in general achieves the second best results based on Table \ref{result_mrr} on the UCI, Wikipedia, Reddit datasets, respectively.  
For the inductive setting, our Ada-DyGNN relatively improves the MRR by 437.0\%, 2.6\%, 20.2\% times over DyRep on the UCI, Wikipedia, Reddit datasets, respectively.  
The dynamic graph neural network models, including  DyRep, TGAT, Jodie, TGN and our Ada-DyGNN, generally achieve better performance than three static models GCN, GraghSAGE and GAT, which demonstrates the necessity of capturing temporal information in dynamic graphs. 
On the Wikipedia and  Reddit datasets,  CTDNE and DyGNN obtain worse results over the static models. We guess that this may be because  CTDNE and DyGNN fail to model the edge features, leading to the information loss.
Table \ref{result_ap} and Table \ref{result_auc} show the experiment results in terms of AP and AUC for the future link prediction task. Our Ada-DyGNN still achieves the best performance over the baselines under all the cases.
Thus, we can come to the same conclusion as the above statement.

\begin{table}
\caption{Ablation study on the three datasets in the transductive setting.}
\label{ablation-tran}
\tabcolsep=4pt
\begin{tabular}{llccc}
                    Datasets & Methods & MRR & AP & AUC \\ \hline
\multirow{2}{*}{UCI}        & Ada-DyGNN                & \textbf{0.0989}  & \textbf{0.9191} & \textbf{0.9259} \\
                            & Ada-DyGNN-agg-w.o.-time  & 0.0827  & 0.9014 & 0.8772 \\ 
                            & Ada-DyGNN-pro-w.o.-time  & 0.0812  & 0.9045 & 0.8997 \\
                            & Ada-DyGNN-select-all     & 0.0742  & 0.8739 & 0.8595 \\
                            & Ada-DyGNN-select-none    & 0.0768  & 0.8318 & 0.8281 \\
                            & Ada-DyGNN-select-random  & 0.0502  & 0.7524 & 0.7354 \\  \hline
\multirow{2}{*}{Wikipedia}  & Ada-DyGNN                & \textbf{0.3323}  & \textbf{0.9898} & \textbf{0.9880} \\
                            & Ada-DyGNN-agg-w.o.-time  & 0.3213  & 0.9816 & 0.9810 \\ 
                            & Ada-DyGNN-pro-w.o.-time  & 0.3224  & 0.9886 & 0.9871 \\
                            & Ada-DyGNN-select-all     & 0.2452  & 0.9763 & 0.9793 \\
                            & Ada-DyGNN-select-none    & 0.3021  & 0.9769 & 0.9745 \\
                            & Ada-DyGNN-select-random  & 0.1447  & 0.9564 & 0.9633 \\  \hline
\multirow{2}{*}{Reddit}     & Ada-DyGNN                & \textbf{0.0568}  & \textbf{0.9936} & \textbf{0.9930} \\
                            & Ada-DyGNN-agg-w.o.-time  & 0.0507  & 0.9871 & 0.9884 \\ 
                            & Ada-DyGNN-pro-w.o.-time  & 0.0479  & 0.9848 & 0.9860 \\
                            & Ada-DyGNN-select-all     & 0.0275  & 0.9758 & 0.9714 \\
                            & Ada-DyGNN-select-none    & 0.0489  & 0.9743 & 0.9695 \\
                            & Ada-DyGNN-select-random  & 0.0218  & 0.9625 & 0.9551 \\  \hline
\end{tabular}

\end{table}

\begin{table}
\caption{Ablation study on the three datasets in the inductive setting.}
\label{ablation-ind}
\tabcolsep=4pt
\begin{tabular}{llccc}
                   Datasets & Methods & MRR & AP & AUC \\ \hline
\multirow{2}{*}{UCI}        & Ada-DyGNN                & \textbf{0.0886}  & \textbf{0.8258}  & \textbf{0.8204} \\
                            & Ada-DyGNN-agg-w.o.-time  & 0.0686  & 0.8089  & 0.7816 \\ 
                            & Ada-DyGNN-pro-w.o.-time  & 0.0711  & 0.8107  & 0.7922 \\
                            & Ada-DyGNN-select-all     & 0.0659  & 0.7842  & 0.7622 \\
                            & Ada-DyGNN-select-none    & 0.0688  & 0.7792  & 0.7596 \\
                            & Ada-DyGNN-select-random  & 0.0458  & 0.7232  & 0.7148 \\  \hline
\multirow{2}{*}{Wikipedia}  & Ada-DyGNN                &\textbf{0.3247}  & \textbf{0.9860}  & \textbf{0.9837} \\
                            & Ada-DyGNN-agg-w.o.-time  & 0.3142  & 0.9842  & 0.9823 \\ 
                            & Ada-DyGNN-pro-w.o.-time  & 0.3171  & 0.9849  & 0.9826 \\
                            & Ada-DyGNN-select-all     & 0.2439  & 0.9834  & 0.9780 \\
                            & Ada-DyGNN-select-none    & 0.2959  & 0.9808  & 0.9786 \\
                            & Ada-DyGNN-select-random  & 0.1490  & 0.9671  & 0.9609 \\  \hline
\multirow{2}{*}{Reddit}     & Ada-DyGNN                & \textbf{0.0981}  & \textbf{0.9880}  & \textbf{0.9868} \\
                            & Ada-DyGNN-agg-w.o.-time  & 0.0942  & 0.9816  & 0.9856 \\ 
                            & Ada-DyGNN-pro-w.o.-time  & 0.0903  & 0.9808  & 0.9847 \\
                            & Ada-DyGNN-select-all     & 0.0529  & 0.9664  & 0.9628 \\
                            & Ada-DyGNN-select-none    & 0.0819  & 0.9737  & 0.9787 \\
                            & Ada-DyGNN-select-random  & 0.0430  & 0.9548  & 0.9578 \\  \hline
\end{tabular}
\end{table}

\begin{figure*}
  \centering
  \includegraphics[width=0.95\linewidth]{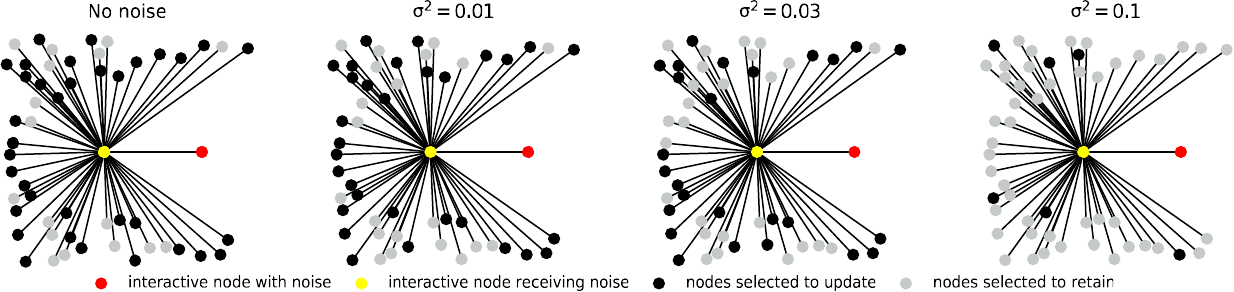}
  \caption{Actions taken by our neighbor selection module under different levels of noise.}
  \label{visualization}
\end{figure*}

\begin{table*}
\centering
\caption{Robustness Analysis of our method in the transductive and inductive settings in terms of MRR on the UCI dataset.}
\label{robustness-mrr}
\begin{tabular}{l|l|c|cc|cc|cc|cc}
\toprule
                    && \multicolumn{1}{c|}{$\sigma^2=0$}   & \multicolumn{2}{c|}{$\sigma^2=0.01$}& \multicolumn{2}{c|}{$\sigma^2=0.03$}& \multicolumn{2}{c|}{$\sigma^2=0.1$}& \multicolumn{2}{c}{$\sigma^2=0.3$}\\
                    \cline{3-11}
                    \multirow{-2}{*}{Models}  & \multirow{-2}{*}{Settings} & MRR& MRR&DEC&MRR &DEC &MRR &DEC & MRR& DEC\\
                    \hline

                            & Transductive &0.0312&0.0311&0.3\%&0.0305&2.2\%&0.0268&14.1\%&0.0152&51.2\%\\\cline{2-11}
\multirow{-2}{*}{DyRep}     & Inductive    &0.0165&0.0164&0.6\%&0.0164&0.6\%&0.0161&2.4\%&0.0126&23.6\%\\  
\hline
                            & Transductive &0.0159&0.0158&0.6\%&0.0156&1.9\%&0.0141&11.3\%&0.0111&30.2\%\\\cline{2-11}
\multirow{-2}{*}{TGN}     & Inductive    &0.0143&0.0142&0.7\%&0.0140&2.1\%&0.0116&18.9\%&0.0081&43.4\%\\ 
\hline
                            & Transductive &0.0989&0.0988&0.1\%&0.0981&0.8\%&0.0958&3.1\%&0.0941&5.0\%\\\cline{2-11}
\multirow{-2}{*}{Ada-DyGNN} & Inductive    &0.0886&0.0885&0.1\%&0.0884&0.2\%&0.0878&0.9\%&0.0855&3.5\%\\
\hline
\end{tabular}
\end{table*}

\begin{table*}
\caption{Robustness Analysis of our method in the transductive and inductive settings in terms of AP on the UCI dataset.}
\centering
\label{robustness-ap}
\begin{tabular}{l|l|c|cc|cc|cc|cc}
\toprule
                    && \multicolumn{1}{c|}{$\sigma^2=0$}   & \multicolumn{2}{c|}{$\sigma^2=0.01$}& \multicolumn{2}{c|}{$\sigma^2=0.03$}& \multicolumn{2}{c|}{$\sigma^2=0.1$}& \multicolumn{2}{c}{$\sigma^2=0.3$}\\
                    \cline{3-11}
                    \multirow{-2}{*}{Models}  & \multirow{-2}{*}{Settings} & AP& AP&DEC&AP &DEC &AP &DEC & AP& DEC\\
                    
\hline
                            & Transductive &0.5507 &0.5448&1.1\% &0.5426&1.5\%&0.5395&2.0\%&0.5273&4.2\%\\\cline{2-11}
\multirow{-2}{*}{DyRep}     & Inductive    &0.5074 &0.4993&1.6\% &0.4990&1.6\%&0.4981&1.8\%&0.4944&2.6\%\\
\hline
                            & Transductive &0.8499 &0.8465&0.4\% &0.8363&1.6\%&0.8273&2.7\%&0.8081&4.9\%\\\cline{2-11}
\multirow{-2}{*}{TGN}       & Inductive    &0.8140 &0.8137&0.1\% &0.8094&0.6\%&0.8064&0.9\%&0.7959&2.2\%\\
\hline
                            & Transductive &0.9191 &0.9154&0.4\% &0.9070&1.3\%&0.9015&1.9\%&0.8830&3.9\%\\\cline{2-11}
\multirow{-2}{*}{Ada-DyGNN}  & Inductive    &0.8258 &0.8256&0.1\% &0.8255&0.1\%&0.8181&0.9\%&0.8157&1.2\%\\
\hline
\end{tabular}
\end{table*}

\begin{table*}
\caption{Robustness Analysis of our method in the transductive and inductive settings in terms of AUC on the UCI dataset.}
\centering
\label{robustness-auc}
\begin{tabular}{l|l|c|cc|cc|cc|cc}
\toprule
                    && \multicolumn{1}{c|}{$\sigma^2=0$}   & \multicolumn{2}{c|}{$\sigma^2=0.01$}& \multicolumn{2}{c|}{$\sigma^2=0.03$}& \multicolumn{2}{c|}{$\sigma^2=0.1$}& \multicolumn{2}{c}{$\sigma^2=0.3$}\\
                    \cline{3-11}
                    \multirow{-2}{*}{Models}  & \multirow{-2}{*}{Settings} & AUC& AUC&DEC&AUC &DEC &AUC &DEC & AUC& DEC\\
                    
\hline
                            & Transductive &0.5720 &0.5706&0.2\% &0.5616&1.8\%&0.5603&2.0\%&0.5436&5.0\%\\\cline{2-11}
\multirow{-2}{*}{DyRep}     & Inductive    &0.4879 &0.4850&0.6\% &0.4781&2.0\%&0.4759&2.4\%&0.4685&4.0\%\\
\hline
                            & Transductive &0.8478 &0.8429&0.6\% &0.8314&1.9\%&0.8262&2.5\%&0.8054&5.0\%\\\cline{2-11}
\multirow{-2}{*}{TGN}       & Inductive    &0.7924 &0.7870&0.7\% &0.7865&0.8\%&0.7805&1.5\%&0.7760&2.1\%\\
\hline
                            & Transductive &0.9259 &0.9237&0.2\% &0.9177&0.9\%&0.9077&2.0\%&0.8880&4.1\%\\\cline{2-11}
\multirow{-2}{*}{Ada-DyGNN}  & Inductive    &0.8204 &0.8170&0.4\% &0.8152&0.6\%&0.8132&0.9\%&0.8110&1.1\%\\
\hline
\end{tabular}
\end{table*}

\subsection{Ablation Study}
In this subsection, we perform ablation study to verify the effectiveness of our two components: 1) time-aware attentional aggregating module which consists of one aggregate process for calculating the interaction message and one information propagation process for calculating the  intermediate embeddings, as shown in Fig. \ref{overall}; 2) the reinforced neighbor selection module. We design two variants for the attention based module, and three variants for the neighbor selection strategies as: 
\begin{itemize}
  \item Ada-DyGNN-agg-w.o.-time: we remove the time-decay coefficient in the aggregate process.
  \item Ada-DyGNN-pro-w.o.-time: we remove the time-decay coefficient in the information propagation process.
  \item Ada-DyGNN-select-all: we update all the neighbors of the interacted nodes.
  \item Ada-DyGNN-select-none: we do not update any neighbors of the interacted nodes.
  \item Ada-DyGNN-select-random: we randomly update the neighbors  of the interacted nodes.
\end{itemize}
{We conduct the ablation study on the three datasets in both transductive and inductive settings. Table \ref{ablation-tran} and Table \ref{ablation-ind} show the experimental results.}
Ada-DyGNN performs better than Ada-DyGNN-agg-w.o.-time and Ada-DyGNN-pro-w.o.-time, indicating that time-related information can boost the performance of our method.
In addition, Ada-DyGNN outperforms the methods using three neighbor selection variants, which demonstrates our robust knowledge adaptation mechanism can effectively determine when to propagate knowledge to other nodes, enabling to learn robust node representations. 

\subsection{Visualizations of Robustness}
In order to intuitively understand our reinforced neighbor selection module,  we design a robustness visualization experiment by showing the actions output by the policy network under different levels of noise added to the UCI dataset. As shown in Fig. \ref{visualization}, the variance $\sigma^2$ of the Gaussian noise is set to $0, 0.01, 0.03$, and  $0.1$ from left to right. The red point is the central node, into which we add the above noise. The noisy information would be blindly propagated to the yellow node and its all neighborhoods if using previous methods. In Fig. \ref{visualization}, we observe that, in the yellow node's neighborhoods, the number of nodes selected to be updated decreases as the level of noise increases when using our method. This indicates that the noisy information could be prevented to some extent by our method, such that the negative influence can be lowered, and thus our method is robust to noise. 

\subsection{Quantitative Robustness Analysis}
We further quantitatively verify the robustness of our model by adding different levels of noise to the UCI dataset. 
In order to demonstrate its robustness, we compare our method with DyRep and TGN, where DyRep achieves the second result in Table \ref{result_mrr} and TGN achieves the second  best result based on Table \ref{result_ap} and \ref{result_auc}. 
After adding the Gaussian noise on each neighbor's embedding, we aggregate the neighborhood information by different methods for updating node embeddings. 
We set the variance $\sigma^2$ of the Gaussian noise to $0.01,0.03,0.1$, and $0.3$ to simulate different levels of noise. 
{Table \ref{robustness-mrr} lists the results in terms of the MRR metric.}
'DEC' denotes the decrements compared to the performance of the corresponding model when $\sigma^2=0$.
As shown in Table \ref{robustness-mrr}, when $\sigma^2=0.01$ and $\sigma^2=0.03$, the noise has a minimal influence on all the models. 
However, when $\sigma^2=0.1$ and $\sigma^2=0.3$, the performances of DyRep and TGN drop dramatically,
while our method has smaller performance drops than them. 
{Table \ref{robustness-ap} and \ref{robustness-auc} show the experimental results in terms of AP and AUC, respectively.
Since the link prediction task evaluated by AP and AUC is  easier than that by the MRR metric, the performances of our method in terms of AP and AUC drop  slightly less  than that in terms of MRR.
Moreover, the decrement of our model is still less than that of the compared methods in both transductive and inductive settings. 
The above robustness analysis once again shows that our method of designing a reinforced knowledge propagation mechanism is robust to noise. 
}

\subsection{Further Study on Attention Mechanism}
In Section \ref{aggregate}, we devise an attention mechanism to aggregate temporal information.
The design of Equation \ref{alpha} for computing attentional coefficient $\alpha$ is based on GAT \cite{velivckovic2017graph}.
However, as pointed out in \cite{brody2021attentive}, there is a static attention problem in GAT, where the learned layer $\mathbf{W}$ and $\mathbf{a}$ are not separated by a non-linearity.
GAT might collapse into a single linear layer, computing a shared ranking of attention coefficients across all nodes, which makes it challenging to distinguish noises.
To this end, GATv2 \cite{brody2021attentive} introduces a straightforward modification to GAT to compute attention, significantly enhancing robustness to noise.
Thus, we conduct another experiment by leveraging the stronger attention mechanism, GATv2,  to investigate the impact of the attention mechanism on the model's performance.
Motivated by GATv2, we modified our Equation \ref{alpha}  as:
\begin{equation}
\label{alpha2}
  \alpha_{si}=\frac{\exp (\mathbf{a}^{\mathrm{T}}\sigma( \mathbf{W}_g[\mathbf{x}_{s}(t)\Vert\phi(\Delta t_i) {\mathbf{x}_{i}(t)}] ))}  {\sum_{j\in \mathbf{\mathcal{N}}_{s}(t)}^{} \exp (\mathbf{a}^{\mathrm{T}}\sigma( \mathbf{W}_g[\mathbf{x}_{s}(t)\Vert\phi(\Delta t_j) \mathbf{x}_{j}(t)] )) }.
\end{equation}

The results are listed in Table \ref{tab:attention}. 
It is evident that the performance of our method can be improved in most cases when applying a stronger attention mechanism.
Thus, it would be interesting to explore other stronger attention mechanisms in future work.
\begin{table}
\caption{Impact of different attention mechanism on the UCI dataset.}
    \label{tab:attention}
    \centering
    \begin{tabular}{ll|ccc}
       &  &MRR &AP &AUC\\ \hline
    \multicolumn{1}{l|}{\multirow{2}{*}{GAT}}& Transductive&0.0989&0.9191&0.9259\\
    \multicolumn{1}{l|}{} &Inductive&0.0886&0.8258&0.8204\\
    \hline
    \multicolumn{1}{l|}{\multirow{2}{*}{GATv2}}& Transductive&0.1072&0.9139&0.9214\\
    \multicolumn{1}{l|}{} &Inductive&0.0903&0.8330&0.8280\\
    \hline
    \end{tabular}    
\end{table}

\subsection{Impact of the Number of Neighbors}
As a trade off between speed and performance, we set a limit to the neighborhood size $k$ in our reinforced based agent. i.e., we only send the most recent $k$ neighbors to the agent for selection.
Thus, we study the impact of different numbers of neighbors on the model performance.
{We search $k$ from \{50, 100, 200\} respectively and test the performance of our method on all the three datasets in both transductive and inductive setting.  Table \ref{Parameter_Analysis-mrr}, \ref{Parameter_Analysis-ap}, and \ref{Parameter_Analysis-auc} report the experimental results in terms of MRR, AP and AUC, respectively.
Note that $k=0$ means that we do not update any neighbor nodes of the two interacted nodes.
We observe that when $k=0$, the performance of the model drops significantly. This illustrates simply dropping all neighbor nodes to avoid noise propagation will lead to significant information loss.
When setting $k$ into 50, 100, or 200, the performance of Ada-DyGNN is relatively stable. 
 Generally, Ada-DyGNN achieves the best performance when $k$ is set to 100 or 200. We set $k=200$ throughout the experiment.}

\begin{table}[ht]
\caption{Impact of the number of neighbors of our method in the transductive and inductive settings in the terms of MRR.}
\centering
\label{Parameter_Analysis-mrr}
\begin{tabular}{ll|cccc}\hline
                                        &      &k=0&k=50&k=100&k=200\\ \hline
\multicolumn{1}{l|}{\multirow{2}{*}{UCI}} &Transductive&0.0768& 0.0981 & 0.0978&0.0989\\
\multicolumn{1}{l|}{}                   &Inductive   &0.0688& 0.0878 & 0.0878&0.0886\\ \hline
\multicolumn{1}{l|}{\multirow{2}{*}{Wikipedia}}&Transductive&0.3021&0.3318&0.3357&0.3323\\
\multicolumn{1}{l|}{}                   &Inductive   &0.2959&0.3210&0.3235&0.3247\\
\hline
\multicolumn{1}{l|}{\multirow{2}{*}{Reddit}}&Transductive&0.0489&0.0531&0.0583&0.0568\\
\multicolumn{1}{l|}{}                   &Inductive   &0.0819&0.0951&0.0971&0.0981 
\\ \hline
\end{tabular}
\end{table}
\begin{table}[ht]
\caption{Impact of the number of neighbors of our method in the transductive and inductive settings in the terms of AP.}
\centering
\label{Parameter_Analysis-ap}
\begin{tabular}{ll|cccc}\hline
                                        &      &k=0&k=50&k=100&k=200\\ \hline
\multicolumn{1}{l|}{\multirow{2}{*}{UCI}} 		&Transductive& 0.8318 & 0.9175 & 0.9175 & 0.9191\\
\multicolumn{1}{l|}{}                     		&Inductive   & 0.7792 & 0.7919 & 0.8283 & 0.8258\\ \hline
\multicolumn{1}{l|}{\multirow{2}{*}{Wikipedia}}		&Transductive& 0.9769 & 0.9886 & 0.9895 & 0.9898\\
\multicolumn{1}{l|}{}                   		&Inductive   & 0.9808 & 0.9856 & 0.9868 & 0.9860\\\hline
\multicolumn{1}{l|}{\multirow{2}{*}{Reddit}}		&Transductive& 0.9743 & 0.9944 & 0.9958 & 0.9936\\
\multicolumn{1}{l|}{}                   		&Inductive   & 0.9737 & 0.9794 & 0.9817 & 0.9880 
\\ \hline
\end{tabular}
\end{table}
\begin{table}
\caption{Impact of the number of neighbors of our method in the transductive and inductive settings in the terms of AUC.}
\centering
\label{Parameter_Analysis-auc}
\begin{tabular}{ll|cccc}\hline
                                        &      &k=0&k=50&k=100&k=200\\ \hline
\multicolumn{1}{l|}{\multirow{2}{*}{UCI}} 		&Transductive& 0.8281 & 0.9144 & 0.9249 & 0.9259\\
\multicolumn{1}{l|}{}                     		&Inductive   & 0.7596 & 0.8093 & 0.8146 & 0.8204\\ \hline
\multicolumn{1}{l|}{\multirow{2}{*}{Wikipedia}}		&Transductive& 0.9745 & 0.9878 & 0.9896 & 0.9880\\
\multicolumn{1}{l|}{}                   		&Inductive   & 0.9786 & 0.9833 & 0.9841 & 0.9837\\\hline
\multicolumn{1}{l|}{\multirow{2}{*}{Reddit}}		&Transductive& 0.9695 & 0.9871 & 0.9933 & 0.9930\\
\multicolumn{1}{l|}{}                   		&Inductive   & 0.9787 & 0.9879 & 0.9867 & 0.9868 
\\ \hline
\end{tabular}
\end{table}

\begin{figure*}
\centering
\subfloat[UCI]
{
\includegraphics[width=0.31\linewidth]
{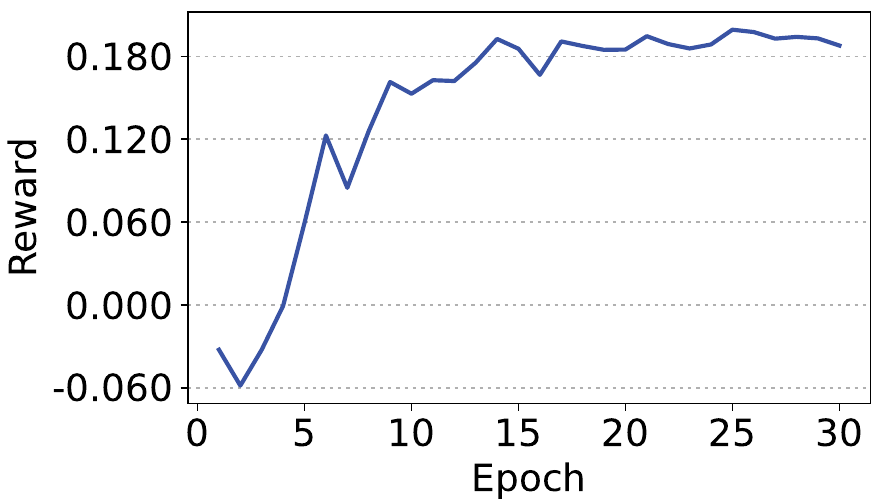}
}
\subfloat[Wikipedia]
{
\includegraphics[width=0.31\linewidth]
{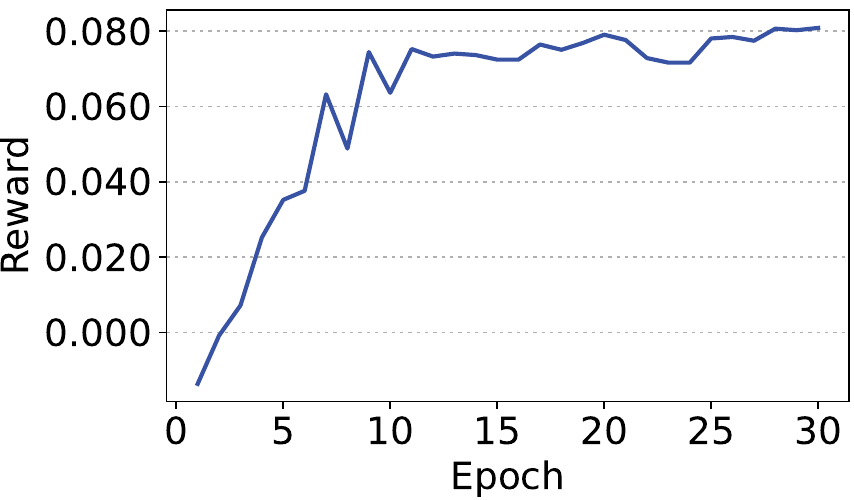}
}
\subfloat[Reddit]
{
\includegraphics[width=0.31\linewidth]
{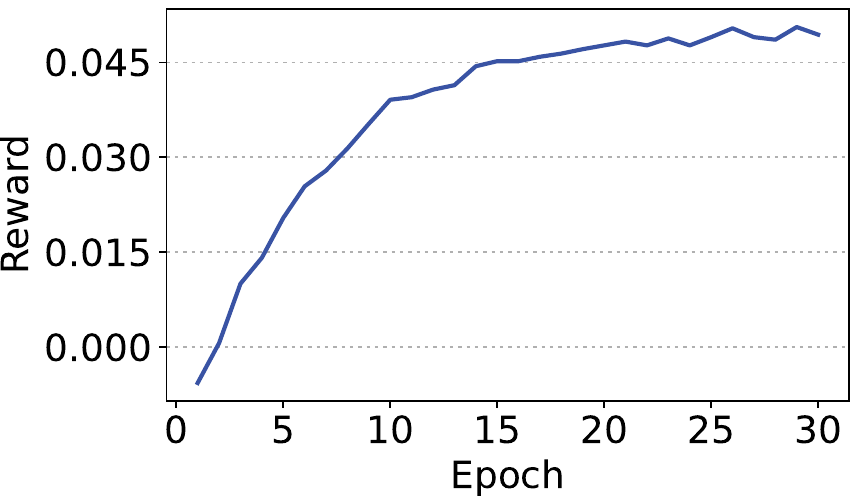}
}
\caption{The reward curves of our method on the UCI, Wikipedia, and Reddit datasets, respectively.}
\label{conv-uci}
\end{figure*}

\begin{figure*}
\centering
\subfloat[UCI]
{
\includegraphics[width=0.31\linewidth]
{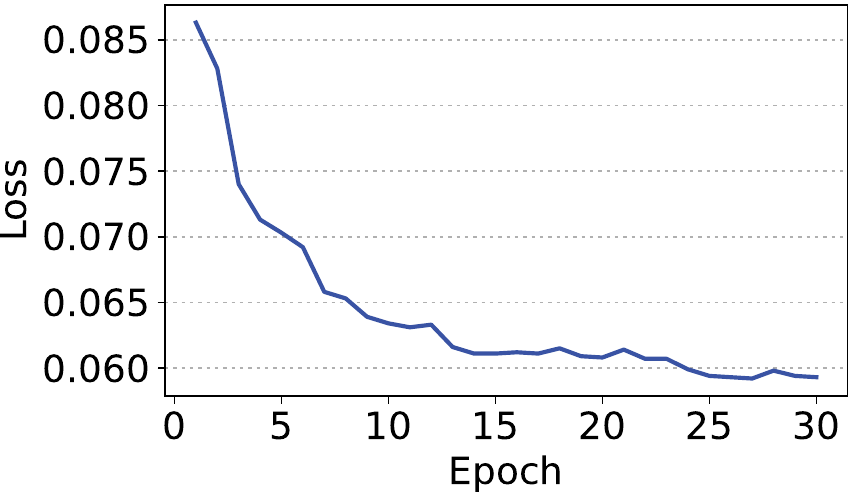}
}
\subfloat[Wikipedia]
{
\includegraphics[width=0.31\linewidth]
{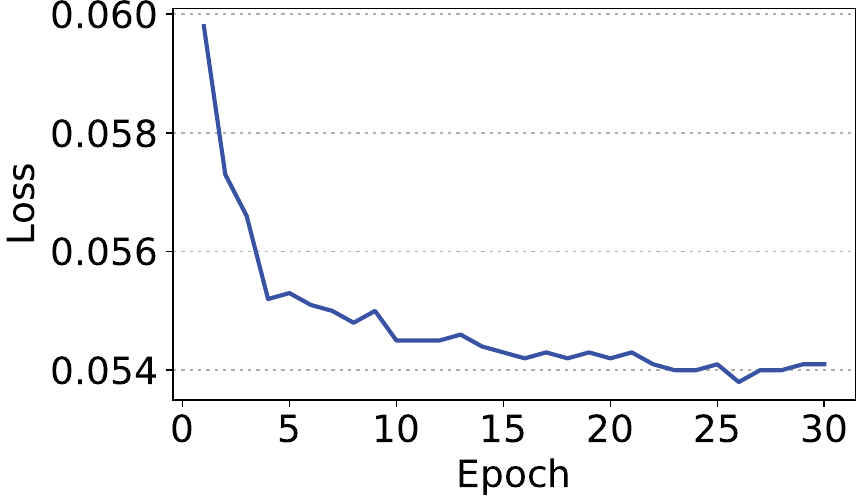}
}
\subfloat[Reddit]
{
\includegraphics[width=0.31\linewidth]
{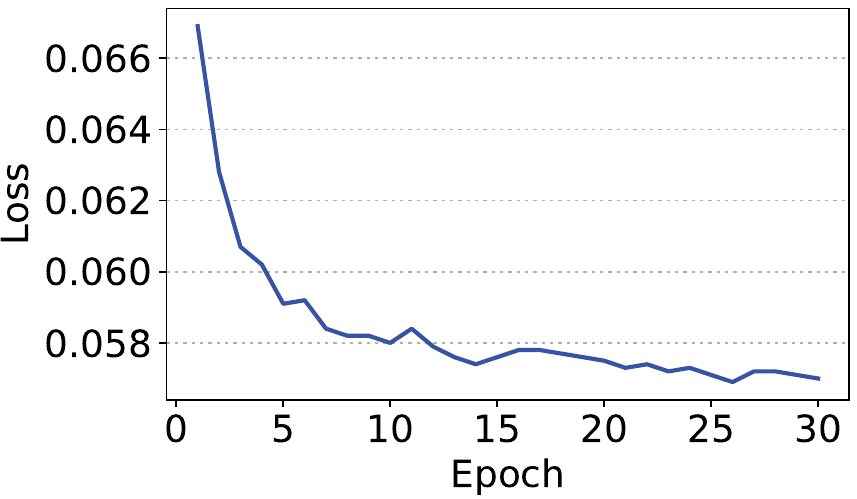}
}
\caption{The loss curves of our method on the UCI, Wikipedia, and Reddit datasets, respectively.}
\label{conv-wiki}
\end{figure*}

\subsection{Impact of Varying Dimensions}
To investigate the impact of varying dimensions on the model's performance, we conduct another experiment using our method with different dimensions on the UCI dataset. Note that in order to minimize the need for hyperparameter tuning, we set $d_n=d_m$ in the experiment. The result are listed in the Table \ref{dimension}. From the table, it is evident that as the dimension increases, the performance is improved. When the dimensions are set to 128, the model exhibits a slight improvement compared to using the dimension 64. 

\begin{table}
\caption{Impact of different dimensions on the UCI dataset.}
    \centering
    \label{dimension}
    \begin{tabular}{ll|ccc}\hline
      &  &dim=32 &dim=64 &dim=128\\ \hline
    \multicolumn{1}{l|}{\multirow{2}{*}{MRR}}& Transductive&0.0832&0.0989&0.1023\\
    \multicolumn{1}{l|}{} &Inductive&0.0726&0.0886&0.0870\\
    \hline
    \multicolumn{1}{l|}{\multirow{2}{*}{AP}}& Transductive&0.9020&0.9191&0.9209\\
    \multicolumn{1}{l|}{} &Inductive&0.8095&0.8258&0.8281\\
    \hline
    \multicolumn{1}{l|}{\multirow{2}{*}{AUC}}& Transductive&0.8978&0.9259&0.9260\\
    \multicolumn{1}{l|}{} &Inductive&0.7970&0.8204&0.8262\\
    \hline
    \end{tabular}
\end{table}
\subsection{Convergence Analysis}
{
Finally, we analyze the convergence of our method. Fig. \ref{conv-uci} and \ref{conv-wiki} show the convergence curves of Ada-DyGNN  on all the three datasets, where the horizontal axis denotes the training epochs and the vertical axis refers to the reward values and the training losses, respectively. Our method converges after around 25 epochs.
}

%% file: sections/conclusion.tex
\section{Conclusion and Future Work}
\label{sec:conclusion}
In this paper, we proposed a robust knowledge propagation  method for dynamic graph neural networks. We devised a reinforcement learning based strategy to dynamically determine whether the embedding of a node should be updated. In this way, we can effectively propagate  knowledge to other nodes and learn robust node representations for dynamic graphs. Extensive experimental results demonstrated that our model outperformed the state-of-the-arts and exhibited  strong robustness against noise.

It is worth noting that, we devise our model based on two important assumptions in dynamic graphs: (1) newer interactions have a greater impact than older ones, and (2) neighbors should exhibit consistency.
These assumptions naturally hold in most of the dynamic graphs. 
For example, in the domain of recommendation systems, where users tend to interact with recently interested items.
However, in case where interaction are irrelevant to history or the graph exhibits inconsistency \cite{liu2020alleviating}, our model might under-perform. 

Several interesting future works can be followed up, including: 
\begin{itemize}
\item Hierarchical decisions: our current Ada-DyGNN model takes actions on the node-wise level, i.e. update or not on a single node. 
In the future work, we could use a hierarchical reinforcement learning strategy to generate decisions at different levels. For example, a higher level decision on the graph restricts the amount of the updating nodes, and a lower level decision decides to update which nodes.  
\item Knowledge adaptation on more complicated graphs: 
there is only one type of relationship between nodes on dynamic graphs in this paper.
We can extend our method to handle multi-relational graph by 
designing multiple strategies to propagated information between neighbor nodes.
\item Diverse graph tasks: we evaluate our proposed model on the future link prediction task in the experiment. We can explore our model on more graph tasks, e.g. node classification, graph classification, community detection. 
\end{itemize}